\documentclass[preprint,12pt]{elsarticle}

\usepackage{amssymb}
\usepackage{amsmath}

\usepackage{fontspec}
\newfontfamily\bengalifont[
  Path = ./,
  Script = Bengali,
  Extension = .ttf
]{NotoSerifBengali-Regular}
\newcommand{\bn}[1]{{\bengalifont #1}}
\usepackage{array}

\newcommand{\bntrans}[2]{%
    \bn{#1}\par\vspace{-2pt}%
    {\tiny\textit{(#2)}}%
}
\usepackage{booktabs}
\usepackage{graphicx}
\usepackage{longtable}
\usepackage{hyperref}
\usepackage[margin=1in]{geometry}
\usepackage{placeins}
\usepackage{multirow}
\usepackage{subcaption}
\hypersetup{
	colorlinks=true,
	citecolor=blue,
	filecolor=black,
	linkcolor=blue
}

\begin{document}

\begin{frontmatter}

\title{IsharaBART: An efficient BART-Based Framework for Low-Resource Bangla Sign Language Translation}

\author{Aonmoy Das}
\ead{aonmoy@std.cu.ac.bd}
\author{Ananna Dev Aishi}
\ead{aishi.eee@std.cu.ac.bd}
\author{Md. Fazlul Kader\corref{cor1}}
\ead{f.kader@cu.ac.bd}
\cortext[cor1]{Corresponding author}
\address{Department of Electrical and Electronic Engineering, University of Chittagong, Chittagong-4331, Bangladesh.}

\begin{abstract}
Millions of Bangla Sign Language (BdSL) users lack reliable automatic translation tools. Existing systems are limited by data scarcity, unidirectional design, and computationally expensive architectures that are not suitable for low-resource deployment. We introduce IsharaBART, a bidirectional generative framework for gloss-sentence translation in BdSL. IsharaBART uses Bidirectional and Auto-Regressive Transformers (BART)-based conditional language modeling with denoising learning objectives. We developed a parallel dataset of 1,000 professionally annotated gloss-sentence pairs, balanced across tenses and structures. We expanded the dataset to 4,831 pairs through four augmentation techniques: rule-based tense transformations, masked token substitutions, back-translation, and one-shot prompting. A compact denoising sequence-to-sequence model was pre-trained from scratch on monolingual Bangla text from six different corpora, then fine-tuned on the annotated and augmented datasets. We compare IsharaBART against four larger baseline models under identical conditions, where it achieves comparable or superior bidirectional performance. On the gloss-to-sentence (G2S) task, it attains Bilingual Evaluation Understudy (BLEU-4)  score up to 51.7, and on the sentence-to-gloss (S2G) task, it achieves 61.40, outperforming multilingual models despite having fewer than 10\% of the parameters. Ablation studies demonstrate that pre-training provides 62.5–90.4\% faster convergence and 11.5-118\% higher BLEU-4 than scratch. Low-rank adaptation (LoRA) reduces trainable parameters by 99.3\%, but with 28.0-52.2\% loss of BLEU-4 compared to full-parameter fine-tuning. IsharaBART establishes a scalable and deployable pipeline for Bangla sign language bidirectional translation, enhancing accessibility in education, healthcare, and social contexts for the deaf and speech-impaired community, while providing a blueprint for similar low-resource languages like Bangla.
\end{abstract}

\begin{keyword}
Sign language translation \sep Bangla sign language \sep Gloss-sentence translation \sep Data augmentation \sep Transformer

\end{keyword}

\end{frontmatter}

\section{Introduction}
\label{sec:introduction}
\subsection{Sign Language and Sign Language Translation}

Sign language stands as one of the most ancient and instinctive method of communication, which serves as the primary mode of communication for people with hearing or speech impairments. According to the World Health Organization, approximately 466 million people worldwide (6.1\% of the global population) experience hearing loss, with projections indicating this number will surpass 900 million by 2050 \citep{who2019}. These sign language users face persistent difficulties in accessing digital services, educational resources, and social participation due to the lack of effective and automatic translation systems. Sign languages are independent linguistic systems with their own lexicons, grammar, and syntax, conveying meaning through manual features such as hand shape, movement, and hand position, together with non-manual features
such as facial expression, mouthing, and eyebrow movement \citep{Guo2025,
DeCoster2024}. Because sign languages rarely have a standardized written form, most technology built for the deaf community must work directly with video or motion data rather than with text, and this is the central technical difficulty that the present work addresses.

SLR methods have progressed from small-vocabulary classifiers trained on hand-crafted features and depth sensors, to deep networks that process body keypoints or raw video using convolutional and recurrent layers, and more recently to transformer-based encoders that better capture long-range temporal structure \citep{Tan2024}. SLT methods build on this progress in two broad ways. Cascading approaches first recognize glosses from a video and then translate the gloss sequence into text, while
end-to-end approaches translate directly from video features to text, sometimes still using glosses as an auxiliary training signal rather than an explicit intermediate step \citep{Tan2024, DeCoster2024}.

\subsection{Bangla Sign Language}
\label{subsec:bdsl}
For Bangla Sign Language (BdSL), which is spoken by the deaf and speech-impaired community in Bangladesh and parts of India, representing one of the world's biggest linguistic regions, the challenges are significantly serious, as technological solutions remain severely underdeveloped compared to other high-resource sign languages such as American Sign Language (ASL) \citep{Othman2012}, Chinese Sign Language (CSL) \citep{Zhou2021} or German Sign Language (DGS) \citep{rwthG2S}. Hearing loss affects a substantial share of the population in the region: recent estimates place significant hearing impairment in Bangladesh at close to 9.6\% of the population, or approximately 13.7 million people \citep{Dhrubo2025}.

BdSL is linguistically distinct from spoken Bangla, with its own sign order and grammatical structure. Therefore, gloss–sentence translation requires cross-representation generation rather than simple word reordering. BdSL is also considerably more low-resource than languages such as ASL, DGS, or CSL. Large parallel corpora with video, gloss, and text annotations, do not yet exist for BdSL, and most prior BdSL work has focused on isolated alphabets or a small set of static signs rather than sentence-level translation \citep{Dhrubo2025}. Most existing BdSL resources are small and recent. Regional dialects and individual differences in signing style add a further layer of variability, with the same word sometimes signed differently across signers and regions. On top of this, there is no widely agreed standard for how BdSL should be glossed, a problem BdSL shares with sign languages in general \citep{DeCoster2024}, which makes it difficult to combine data from different sources or compare results across studies.

\subsection{Motivation and Research Gap}
BdSL recognition and translation research has mostly been in very resource limited settings. Most of the past recognition works have focused on individual sign gestures of words, vowels and consonants, and numbers \citep{Hossen2018, Islam2019, Siddique2023, Raihan2024}. These works have been highly effective on isolated sign recognition, but their approach do not address the recognition and translation of continuous sign language gestures. Recent works have started to address continuous translation \citep{Das2025, Dhrubo2025}. These studies are limited to small and shallow architectures with limited capacity to generalize. The datasets used in these studies are limited by very small vocabularies. Moreover, a fundamental requirement for building effective sign language communication tools is bidirectional translation. Past works do not address bidirectional translation and evaluation. To the best of our knowledge, these systems do not utilize modern pre-trained language models or use architectures that could enable both G2S and sentence-to-gloss (S2G) generation.

These research gaps come from three fundamental challenges in low resource languages like BdSL. First, BdSL does not yet have large scale gloss-sentence pair parallel dataset. Such high quality dataset are essential to training high performing models. Second, existing studies use large, resource heavy multilingual models. These models may not have optimal tokenization or scoped abilities specific to BdSL gloss-sentence translation. Third, real-world application of translation tools require small, compact models that are capable of running on edge devices and mobile platforms with limited computational resources.

\subsection{Contributions of This Study}
In this work, we introduce IsharaBART, a bidirectional generative framework specifically designed for BdSL gloss-sentence translation that addresses the aforementioned fundamental gaps through these key contributions:
\begin{enumerate}
    \item We developed IsharaBART, a small (56M) denoising seq2seq model pre-trained on monolingual Bangla text from six different corpora. IsharaBART achieved competitive bidirectional gloss-sentence translation performance while using substantially fewer parameters (78-90\% fewer) than the baseline models. 
     \item We developed a parallel gloss-sentence dataset for BdSL, which contains 1,000 professionally annotated gloss-sentence pairs. The data is balanced across tenses and syntactic structures. The dataset also contains 3,831 synthetically augmented pairs using four complementary augmentation techniques.
     \item We proposed a comprehensive framework for sign language gloss-sentence data augmentation and data validation that expands our 1,000-pair dataset to 4,831 pairs while preserving structural, semantic, and lexical consistency.
      \item We conducted a detailed bidirectional evaluation of IsharaBART by comparing against four baseline models and different architectures under controlled experimental conditions (identical datasets, training protocols, and evaluation metrics).  It is demonstrated that monolingual pre-training enables dramatic convergence speedups (62.5-90.4\% faster) in gloss-sentence translation tasks.    
    \item We performed ablation studies comparing full fine-tuning against Low-rank adaptation (LoRA) to provide relevant information for deployment scenarios with different computational resources.
\end{enumerate}

\subsection{Organization of This Paper}
The remainder of this paper is structured in the following way. Section \ref{sec:literature_review} provides a detailed literature review in general sign language and BdSL translation. Section \ref{sec:methodology} describes the methodology used in the study. Section \ref{sec:experiments} describes the experimental design and evaluation metrics. Section \ref{sec:results} reports and provides analysis of the results from the experiments and ablation studies. Section \ref{sec:discussion} further discusses the findings of the study and provides a more abstract implications of those results. Finally, Section \ref{sec:conclusion} outlines the limitations of the study and provides potential future research directions to address those limitations.

\section{Literature review}\label{sec:literature_review}
Research on automatic sign language processing is broadly categorized into two different tasks \citep{DeCoster2024}. Sign Language Recognition (SLR) identifies signs in a video. It usually outputs these gestures as glosses or words in the order they appear in the video. Sign Language Translation (SLT) takes these sequence of glosses and outputs meaningful sentences to complete communication \citep{DeCoster2024}. The reverse direction of generating gesture sequences from spoken texts also falls under this category.

\subsection{SLR Evolution}
\label{subsubsec:general_evolution}
SLR has progressed from classification of gestures by simpler hand centric feature extraction to more complex and deeper multimodal deep learning \citep{Zhang2024}. Early works used depth-based CNNs for fingerspelling \citep{Kang2015}, CNNs for 3D pose estimation of hands \citep{Oberweger2015}, RGB-depth-skeleton fusion \citep{Huang2015}, and restricted Boltzmann machines for fusing multidimentional features \citep{Rastgoo2018}. Later studies used 3D residual CNNs and BiLSTMs for continuous sign recognition \citep{Wei2019}. Hasanov et al. studied isolated Azerbaijani fingerspelling \citep{Hasanov2023}. Aiswarya et al. used keyframe selection to process continuous sign videos \citep{Aiswarya2024}. Hao et al. combined sensing-based CNNs with BiLSTM-based attention \citep{Hao2024}. Renjith et al. used Transformers to combine motion and visual features \citep{Renjith2026}. Barrister et al. used multiscale Transformer and convolutional features to detect Mizo signs \citep{Barrister2026}. Wu et al. added sign cues to trimodal speech processing \citep{Wu2024}. Talaat et al. developed a sign-language translation system that generates Arabic avatars in real time \citep{Talaat2024}. However, most studies focus on isolated-sign recognition or specific applications. Few address translation in both directions between sign-language glosses and sentences.

\subsection{Transformer-Based SLT}
\label{subsubsec:transformer_based_slt}
Researchers moved from CNN-LSTM models to transformer-based models for continuous SLR and translation. Camgoz et al. \citep{Camgoz2020} used transformers for end-to-end sign language translation. They tested their model on the RWTH-PHOENIX-Weather-2014T dataset. The dataset included 1,066 unique signs from 9 signers in unconstrained settings. Their model performed better than earlier gloss-to-text methods. It improved the BLEU-4 score from 9.58 to 21.80. Chaudhary et al. \citep{Chaudhary2023} proposed SignNet II. It uses a transformer-based architecture for two-way sign language translation. The model used a dual learning mechanism to train sign-to-text (SLT) and text-to-sign (TSL) networks together.  This approach performed better than singly-trained networks and achieved noteworthy improvement in BLEU-1 to BLEU-4 scores in both translation directions. Hao et al. \citep{Hao2025} proposed a Convolutional Vision Transformer (CvT) model combining hierarchical convolutional tokenization with transformer self-attention mechanisms. They evaluated the model on two publicly available sign language datasets; a digit dataset (1,712 images, 10 classes) and an alphabet-symbol dataset (87,000 images, 29 classes). The CvT model achieved 99\% accuracy, outperforming traditional CNNs and competing vision transformer (ViT) models. In \citep{Wesley2025}, the authors proposed a transformer-based SLT system that relies exclusively on 2D body keypoints extracted by MediaPipe Holistic instead of raw video frames or pixel embeddings. They introduced spatial–temporal augmentation on keypoints, convolutional embeddings and CTC loss with positional encoding for Sign2Gloss. They also fine-tuned BART for Gloss2Text. On Phoenix14T, they achieved competitive gloss-based BLEU-4 of 21.30 and gloss-free BLEU-4 of 15.70 (test set), while on How2Sign the gloss-free model reached only 3.87 BLEU-4. Although the keypoint-only pipeline significantly reduces computational cost and improves deployability, performance remains limited by the narrow weather-news domain of the datasets. Jin et al. \citep{Jin2025} proposed a sign language production (SLP) method that employs Locality Sensitive Hashing-Clustering to automatically group similar embedded word vectors. This helps the model capture long-distance dependencies in text-to-pose sequence generation. They also used multi-scale convolutional feature extraction module to capture local gloss features. Their transformer-based architecture was evaluated on the PHOENIX14T dataset and demonstrated improvements over baseline SLP methods. However, the model is restricted to a single high-resource dataset and focuses solely on the production direction, leaving the inverse G2S translation task and low-resource sign languages entirely unaddressed.

\subsection{BdSL-Specific Research and Challenges}
\label{subsubsec:bdsl_specific}
Research on BdSL recognition has progressed from CNN-based isolated alphabets to transformer-based real-time translation systems, establishing critical datasets and benchmarks for the Bangladeshi Deaf community. Authors in \citep{Islam2019} trained a CNN on the Ishara-Bachon digit dataset, achieving ~95\% accuracy for classifying BdSL digits. This model was among the first deep-learning models for BdSL and used grayscale digit images. The primary limitation was that it was tested only on static, isolated digit images from a small dataset. Hossen et al. \citep{Hossen2018} developed a transfer-learning deep convolutional neural network (DCNN) to recognize 37 static BdSL alphabet signs. Using 1,147 images of 31 signers, fine-tuning a pre-trained network yielded 96.3\% training accuracy but only 84.7\% on validation. The model showed signs of overfitting, as indicated by the substantial difference between its training and validation accuracies. In \citep{Karim2023}, the researchers applied transfer learning with YOLOv5 for BdSL recognition. They extracted frames from a large BdSL video collection and annotated 24.2 GB of data. Using pre-trained YOLOv5, the system achieved~91.6\% average classification accuracy, but the performance depends on video frame quality and annotation density. Deep detectors were deployed on an edge device in \citep{Siddique2023}. Using the Okkhornama BdSL image set (3,760 images, 49 classes), their YOLOv7-Tiny model achieved real-time inference on Nano. The limitation was that the dataset was small, and modest mAP meant many classes were still missed. Raihan et al. \citep{Raihan2024} proposed a CNN with squeeze-and-excitation (CNN–SE) for BdSL recognition. They trained on the KU-BdSL alphabet dataset (1,500 images of 30 signs) and incorporated a smartphone app. This model achieved 99.86\% test accuracy on KU-BdSL. The near-perfect accuracy likely reflects limited variability in KU-BdSL, real-world performance might drop under varied lighting or new signers. The authors in \citep{Das2025} constructed a custom BdSL dataset and proposed a deep LSTM-based BdSL translator that modeled temporal sign sequences using stacked LSTM layers without relying on gloss-level supervision. Their system achieved strong translation accuracy on constrained sentence vocabularies. However, the approach is limited by scalability issues inherent to LSTMs and does not leverage large pre-trained language models or Transformer-based attention mechanisms. In \citep{Dhrubo2025}, researchers developed a real-time sign-to-text BdSL translator. They curated a 102-word dataset with 1,200 sentences and extracted hand/body keypoints. A custom Transformer model (with efficient landmark encoding) mapped sign sequences to Bangla text, reportedly achieving the highest accuracy to date on this benchmark. Key limitations are that the system uses a small, fixed vocabulary of everyday words and relies on robust landmark tracking.

\begin{table*}[t]
\centering
\caption{Evolution of SLR  and SLT: from hand-centric to transformer-based approaches.}
\resizebox{\textwidth}{!}{
\begin{tabular}{@{}llllp{5.5cm}p{5cm}@{}}
\toprule
\textbf{Study} & \textbf{Year} & \textbf{Approach} & \textbf{Dataset/Language} & \textbf{Key Achievement} & \textbf{Limitations} \\ 
\midrule
\multicolumn{6}{l}{\textit{\textbf{Phase I: Hand-Centric \& Multi-Modal Foundations (2015-2019)}}} \\
\midrule
Kang et al. \citep{Kang2015} & 2015 & CNN + Depth Maps & 31 ASL alphabets/digits & 99.99\% (same signer), 83.58-85.49\% (cross-subject) & Isolated gestures only; no temporal continuity \\
Oberweger et al. \citep{Oberweger2015} & 2015 & CNN for 3D Joints & Depth-based hand pose & 10-20mm mean joint error & Requires depth calibration; fails with occlusions \\
Huang et al. \citep{Huang2015} & 2015 & 3D CNN Multi-Modal & RGB-Depth-Skeleton & 94.2\% accuracy; first multi-modal fusion & Static feature fusion; limited to short sequences \\
Rastgoo et al. \citep{Rastgoo2018} & 2018 & RBM Fusion & Massey/NYU/ASL datasets & 11-28\% improvement over prior work & Superseded by LSTM-based methods \\
Wei et al. \citep{Wei2019} & 2019 & 3D ResNet + BiLSTM & CSL & 95.2\% precision; sentence-level translation & Rule-based grammar; limited scalability \\
\midrule
\multicolumn{6}{l}{\textit{\textbf{Phase II: Transformer Revolution (2020-2023)}}} \\
\midrule
Camgöz et al. \citep{Camgoz2020} & 2020 & Sign Transformers & PHOENIX-2014T (1,066 signs) & 21.80 BLEU-4 vs. 9.58 prior SOTA & Gloss-dependent; computationally intensive \\
Chaudhary et al. \citep{Chaudhary2023} & 2023 & SignNet II (Dual) & Bidirectional SLT-TSL & Outperforms unidirectional on all BLEU metrics & Requires parallel sign-text training data \\
Hao et al. \citep{Hao2025} & 2025 & CVT & SL digits (10 classes), alphabets (29 classes) & 99\% accuracy; beats CNNs and ViTs & Tested on isolated static images only \\
\midrule
\multicolumn{6}{l}{\textit{\textbf{Phase III: BdSL-Specific Research (2018-2025)}}} \\
\midrule
Hossen et al. \citep{Hossen2018} & 2018 & Transfer Learning DCNN & 1,147 images, 31 signers (BdSL) & 96.3\% train, 84.7\% validation & Severe overfitting; small dataset \\
Islam et al. \citep{Islam2019} & 2019 & CNN & Ishara-Bachon digits & ~95\% digit classification & Grayscale only; no dynamic gestures \\
Karim et al. \citep{Karim2023} & 2023 & YOLOv5 Transfer & 24.2GB BdSL video frames & ~91.6\% average accuracy & Quality-dependent; annotation-intensive \\
Siddique et al. \citep{Siddique2023} & 2023 & YOLOv7-Tiny Edge & Okkhornama (3,760 images, 49 classes) & Real-time on Jetson Nano & Low mAP; many missed classes \\
Raihan et al. \citep{Raihan2024} & 2024 & CNN-SE & KU-BdSL (1,500 images, 30 signs) & 99.86\% test accuracy & Dataset overfitting; limited variability \\
Das et al. \citep{Das2025} & 2025 & Stacked LSTM & BdSL words (100 words, 500 videos) & 89\% test accuracy & LSTM scalability; no large language model (LLM) integration \\
Dhrubo et al. \citep{Dhrubo2025} & 2025 & Transformer + Landmarks & 102 words, 1,200 sentences (BdSL) & Highest reported BdSL accuracy & Fixed vocabulary; landmark-dependent \\
\bottomrule

\end{tabular}
}
\label{tab_SLR}
\end{table*}

The literature on SLT shows a clear trajectory as depicted in Table~\ref{tab_SLR} from isolated sensor-based recognition and small static datasets toward transformer-based end-to-end translation, but critical gaps remain for low-resource languages. Especially, the work on BdSL has largely been confined to small, isolated image/video datasets. Most studies focus on sign-to-text translation in one direction. They do not use large pre-trained language models to support both gloss-to-sentence and sentence-to-gloss translation. Some studies use simple models to process sign sequences. Others use large multilingual models. These models may not work well on devices with limited resources.

Our work addresses these problems. We develop a bidirectional generative framework for Bangla Sign Language (BdSL). We use a Bidirectional and Autoregressive Transformer (BART) model and we pre-train the model on multiple large Bangla text corpus. The model is then fine-tuned for both G2S and S2G translation direction. For pre-training we use conditional language modeling with specific denoising learning objectives. Our approach goes further than simple sequence mapping and uses training method that is very similar to the structural properties of BdSL. Our approach also attemps to reduce vocabulary limits and overfitting in earlier LSTM and transformer models. We describe each step, from data preparation to model training and evaluation. Our primary goal is to build a lightweight system that can produce gloss-sentence translation with high quality in both directions.

\section{Methodology}\label{sec:methodology}
In this section, we discuss the proposed methodology for the successful development of a language model for a bidirectional gloss-sentence translation system targeted towards BdSL. The proposed methodology to research and develop the primary language model responsible for sign language gloss to sentence synthesis is depicted in Figure \ref{fig:methodology1}, and a step-by-step description of the methodology is provided in the following subsections.

\subsection{Data Acquisition}
\subsubsection{Pre-training Dataset}
\label{subsubsec:pretrain_data}

The pre-training corpus was assembled from six complementary Bangla text collections: Sangraha \citep{sangraha}, {OSCAR (Bangla) \citep{oscar}, C4 (Bangla) \citep{c4}, {Bangla Wikipedia \citep{wikidump}, Bangla books review collection \citep{banglabook}, and a Bangla dialogues \citep{dialogues} dataset. These sources were selected to provide broad coverage of written and conversational registers, genres, and domain styles necessary for robust language modeling in downstream sign-language tasks. To maintain efficient I/O for large-scale training, each cleaned dataset was tokenized and serialized into contiguous binary files. Separate files were produced for training and validation splits to enable streaming consumption by standard pre-training toolchains while preserving document boundaries. All corpora were tokenized with the trained tokenizer prior to serialization. The tokenized corpora are stored as contiguous binary streams (one stream per dataset per split) to support deterministic, high-throughput sampling during pre-training. A SentencePiece byte pair encoding tokenizer was trained from scratch on a 60 GB Bangla corpus. The resulting tokenizer has a vocabulary of 52,001 subwords, averaging 1.33 tokens per word (127.5\% of ideal tokenization efficiency) and a character-to-token compression ratio of 2.96.

\begin{figure*}[t]
    \centering
    \includegraphics[width=\textwidth]{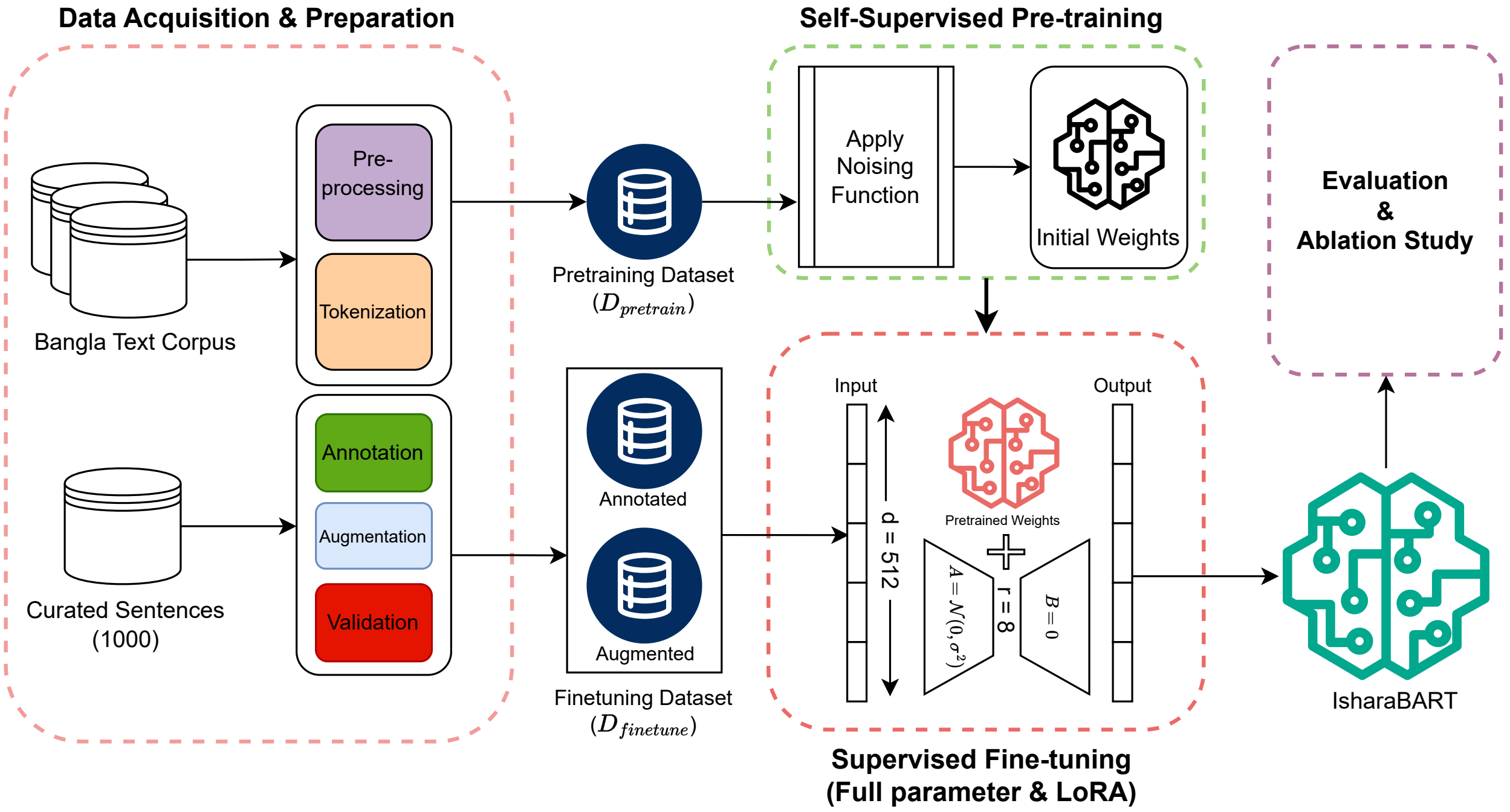}
    \caption{Proposed methodology for developing IsharaBART.}
    \label{fig:methodology1}
\end{figure*}

\subsubsection{Fine-tuning Dataset}
\label{subsubsec:finetune_data}
We created the fine-tuning dataset for G2S and S2G tasks in BdSL. First, we used Sketch Engine \citep{sketchengine} and the bntenten21 corpus to find common Bangla words. We analyzed word frequency and part-of-speech (POS) labels. We ranked the words by frequency and checked their POS labels. We then selected 100 words. The list includes verbs (30\%), adjectives (20\%), nouns (20\%), adverbs (10\%), pronouns (10\%), and conjunctions (10\%).

We wrote ten sentences for each word. These sentences covered present, past, and future tenses in a 4:3:3 ratio. We also included different sentence structures. A professional BdSL interpreter reviewed all sentence pairs. We mainly used the BanglaTense \citep{banglatense} and BanglaSarc \citep{banglasarc} datasets to collect sentences. We modified some sentences to keep their grammar consistent and make them suitable for gloss annotation.

\subsection{Data Augmentation}
\label{subsec:augmentation}

To address the lack of data in BdSL research, we used four complementary augmentation techniques to expand our annotated dataset of 1,000 gloss-sentence pairs. These methods were designed to introduce lexical, syntactic, and semantic variations while preserving the core G2S mappings essential for sequence-to-sequence learning. The augmented data was manually filtered for grammatical correctness and semantic coherence, which resulted in an augmented dataset size of 3,831 pairs and a total of 4831 pairs (annotated + augmented). The final dataset contains 1,000 annotated pairs (20.7\%), 951 pairs from tense transformation (19.7\%), 881 pairs from masked token substitution (18.2\%), 999 pairs from back-translation (20.7\%), and 1,000 pairs from one-shot prompting (20.7\%). Each method contributes a similar number of pairs.

We use four methods to expand the annotated dataset. These methods are based on existing data augmentation techniques. We describe our changes below.

\textbf{Rule-based tense transformation} changes the tense of each Bangla sentence using predefined grammar rules. For example, it can change a present-tense sentence to future tense. We also update the matching gloss tokens to keep both sides aligned. Previous studies used rules to generate gloss data for sign language translation \citep{Othman2012, Moryossef2021}. However, they generated glosses from monolingual spoken text. Our method starts with an existing aligned sentence and gloss pair. It changes both sides of the pair. We designed the tense rules specifically for this study. \textbf{Masked token substitution} masks one content word (noun, verb, or adjective) in each sentence and replaces it with the top grammatically correct prediction of BanglaBERT \citep{BanglaBERT}, updating the corresponding gloss token. This follows contextual augmentation, where words are replaced by those a language model predicts from context \citep{Kobayashi2018}, and its use on gloss data with a masked language model \citep{Jang2022}. \textbf{Back-translation paraphrasing} translates each sentence to English with NLLB-200 \citep{NLLB2022} (1.3B parameters) and back to Bangla with IndicTrans2 \citep{IndicTrans2} (1B parameters), and keeps the original gloss. Back-translation is a standard augmentation method for low-resource translation \citep{Sennrich2016}, and round-trip paraphrases paired with the original gloss have been used for Korean Sign Language \citep{Jang2022}. Back-translation that requires a trained gloss-to-text model is unreliable when gloss data is scarce \citep{Moryossef2021}. Our round trip only uses pre-trained text translation models, so it needs no gloss data. Failed translations and paraphrases that changed the meaning were removed during manual filtering. \textbf{One-shot prompt-based synthesis} uses Llama 3.3-70B \citep{Llama3} to generate new gloss-sentence pairs. Each prompt contains a single annotated pair and asks for a new pair with altered content and a similar grammatical style (Table~\ref{tab:oneshot-model}). Prompting an LLM with seed examples to create new training data follows \citep{Wang2023} and has been used to augment data for low-resource languages \citep{Whitehouse2023}. Ours is a minimal one-example variant. We use a 70B model because this kind of generation is expected to work better with larger models \citep{Wang2023}. Since such outputs can contain invalid examples \citep{Wang2023, Whitehouse2023}, all generated pairs were parsed and validated for BdSL compatibility. Table~\ref{tab:augmentation-examples} provides illustrative examples from each technique.

\begin{table}[t]
\centering
\caption{Model and prompt template used for one-shot prompting augmentation.}
\label{tab:oneshot-model}

\begin{tabular}{p{0.28\columnwidth} p{0.20\columnwidth} p{0.42\columnwidth}}
\toprule
\textbf{Model} & \textbf{Provider} & \textbf{Details} \\
\midrule
llama-3.3-70b-versatile & Groq API &
70B parameters; temperature=0.85 for diversity; max\_tokens=150; top\_p=1 \\
\midrule

\multicolumn{3}{c}{\textbf{Prompt Template}} \\
\midrule

\multicolumn{3}{p{0.90\columnwidth}}{%
I will provide you an example of a BdSL Gloss and its corresponding Bangla Sentence.\par\vspace{0.5em}
Example Input:\newline
Sentence: \{Example Sentence\}\newline
Gloss: \{Example Gloss\}\par\vspace{0.5em}
Task: Generate a completely NEW Gloss and Sentence pair.\par\vspace{0.5em}
1. Structure: Follow the same grammatical style as the example.\newline
2. Content: The meaning MUST be different (change the subject, object, or verb).\newline
3. Format: Output ONLY the result in this exact format:\newline
\quad Sentence: [New Sentence]\newline
\quad Gloss: [New Gloss]
} \\

\bottomrule
\end{tabular}
\end{table}

\begin{table}[!t]
\centering
\small
\caption{Examples of the augmentation strategies used to generate additional gloss--sentence pairs. Colored text marks the token substituted by each technique (red: original, blue: replacement).}
\label{tab:augmentation-examples}

\begin{tabular}{p{0.22\columnwidth} p{0.73\columnwidth}}
\toprule
\textbf{Technique} & \textbf{Step-by-Step Process and Example} \\
\midrule

\textbf{Rule-Based Transformation} &
\textbf{Original:}\newline
Sentence: \bn{তুমি তাড়াতাড়ি কাজটি \textcolor{red}{করো}।}\newline
Gloss: \bn{তুমি তাড়াতাড়ি কাজ এটি \textcolor{red}{কর}}\newline
\par\vspace{0.3em}
\textbf{Step 1:} Identify verb tense/mood (\bn{\textcolor{red}{করো}} - imperative present)\newline
\textbf{Step 2:} Apply tense transformation rule (imperative $\rightarrow$ future)\newline
\textbf{Step 3:} Transform verb: \bn{\textcolor{red}{করো}} $\rightarrow$ \bn{\textcolor{blue}{করবে}}\newline
\textbf{Step 4:} Update gloss accordingly: \bn{\textcolor{red}{কর}} $\rightarrow$ \bn{\textcolor{blue}{করা হবে}}
\par\vspace{0.3em}
\textbf{Augmented:}\newline
Sentence: \bn{তুমি তাড়াতাড়ি কাজটি \textcolor{blue}{করবে}।}\newline
Gloss: \bn{তুমি তাড়াতাড়ি কাজ এটি \textcolor{blue}{করা হবে}}
\\
\midrule

\textbf{Masked Token Substitution} &
\textbf{Original:}\newline
Sentence: \bn{সে পরিশ্রমে ভাল \textcolor{red}{রেজাল্ট} করেছে।}\newline
Gloss: \bn{সে কাজ পরিশ্রম অনেক ভাল \textcolor{red}{রেজাল্ট} পাওয়া}
\par\vspace{0.3em}
\textbf{Step 1:} Randomly select a content word (\bn{\textcolor{red}{রেজাল্ট}})\newline
\textbf{Step 2:} Mask the token: \bn{সে পরিশ্রমে ভাল} [MASK] \bn{করেছে।}\newline
\textbf{Step 3:} Use BanglaBERT to predict: [MASK] $\rightarrow$ \bn{\textcolor{blue}{ফল}}\newline
\textbf{Step 4:} Substitute in sentence and update gloss
\par\vspace{0.3em}
\textbf{Augmented:}\newline
Sentence: \bn{সে পরিশ্রমে ভাল \textcolor{blue}{ফল} করেছে।}\newline
Gloss: \bn{সে কাজ পরিশ্রম ভাল \textcolor{blue}{ফল} পাওয়া}
\\
\midrule

\textbf{Back-Translation} &
\textbf{Original:}\newline
Sentence: \bn{\textcolor{red}{তার সাথে আমার প্রায়ই দেখা হয়।}}
\par\vspace{0.3em}
\textbf{Step 1:} Translate to English using NLLB200:\newline
``I see him often.''
\newline
\textbf{Step 2:} Translate back to Bangla using IndicTrans2:\newline
\bn{\textcolor{blue}{আমি প্রায়ই তাকে দেখি।}}\newline
\textbf{Step 3:} Generate gloss for paraphrased sentence
\par\vspace{0.3em}
\textbf{Augmented:}\newline
Sentence: \bn{\textcolor{blue}{আমি প্রায়ই তাকে দেখি।}} (paraphrased)\newline
Gloss: \bn{\textcolor{blue}{তার সাথে আমার হঠাৎ হঠাৎ দেখা হয়}}
\\
\midrule

\textbf{One-Shot Prompt} &
\textbf{Seed Example:}\newline
Sentence: \bn{\textcolor{red}{তুমি তাড়াতাড়ি কাজটি করো}।}\newline
Gloss: \bn{\textcolor{red}{তুমি তাড়াতাড়ি কাজ} এটি \textcolor{red}{কর}}
\par\vspace{0.3em}
\textbf{Step 1:} Provide seed pair to LLM (Llama-3.3-70B)\newline
\textbf{Step 2:} Prompt LLM\newline
\textbf{Step 3:} LLM generates by changing subject/object/verb
\par\vspace{0.3em}
\textbf{Generated:}\newline
Sentence: \bn{\textcolor{blue}{সে ধীরে ধীরে খাবারটি খায়}।}\newline
Gloss: \bn{\textcolor{blue}{সে ধীরে ধীরে খাবার} এটি \textcolor{blue}{খা}}
\\

\bottomrule
\end{tabular}
\end{table}

\FloatBarrier

\subsection{Dataset Validation}
\label{subsec:validation}

We first evaluate the quality of the fine-tuning dataset through statistical analysis. The parameters we mainly analyse are sentence length, gloss-to-text length ratios, semantic embeddings, and lexical diversity.

The annotated and augmented datasets have very similar lengths. The mean number of words in the annotated and augmented gloss-sentence samples are 5.45 ($SD=1.37$) and 5.51 ($SD=1.55$) respectively (Figure~\ref{fig:sentence_length}). Kernel density estimation (KDE) \citep{KDE} shows that both datasets have similar gloss-to-text length ratio (GTR). The GTR distribution peaks at about 1.0 (Figure~\ref{fig:gtr_distribution}). We also project the BanglaBERT embeddings \citep{BanglaBERT} of the samples in both dataset into two dimensions using UMAP \citep{UMAP}. The annotated and augmented samples show a centroid separation of 1.04 units. The UMAP also shows an overlap of 25.0\% (Figure~\ref{fig:umap_semantic}). These indicates small overlapping but distinct semantic characteristics. Heaps' Law \citep{HeapsLaw}, $V=K\cdot N^{\beta}$, shows vocabulary growth. The estimated $\beta$ values are 0.622 for the annotated dataset (vocabulary: 1,027), 0.575 for the augmented dataset (1,859), and 0.553 for the combined dataset (Figure~\ref{fig:heaps_law}).

\begin{figure*}[t]
    \centering

    \begin{subfigure}{\textwidth}
        \centering
        \includegraphics[width=\textwidth]{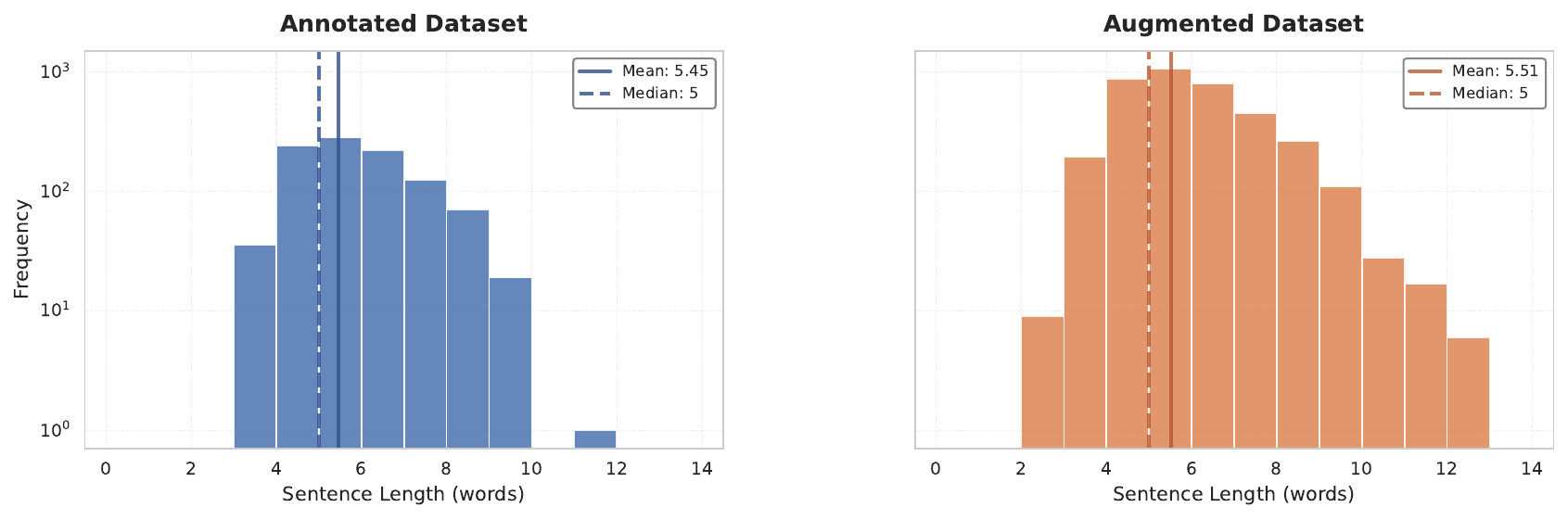}
        \caption{Sentence length distributions.}
        \label{fig:sentence_length}
    \end{subfigure}

    \vspace{3pt}

    \begin{subfigure}[t]{0.32\textwidth}
        \centering
        \includegraphics[width=\linewidth]{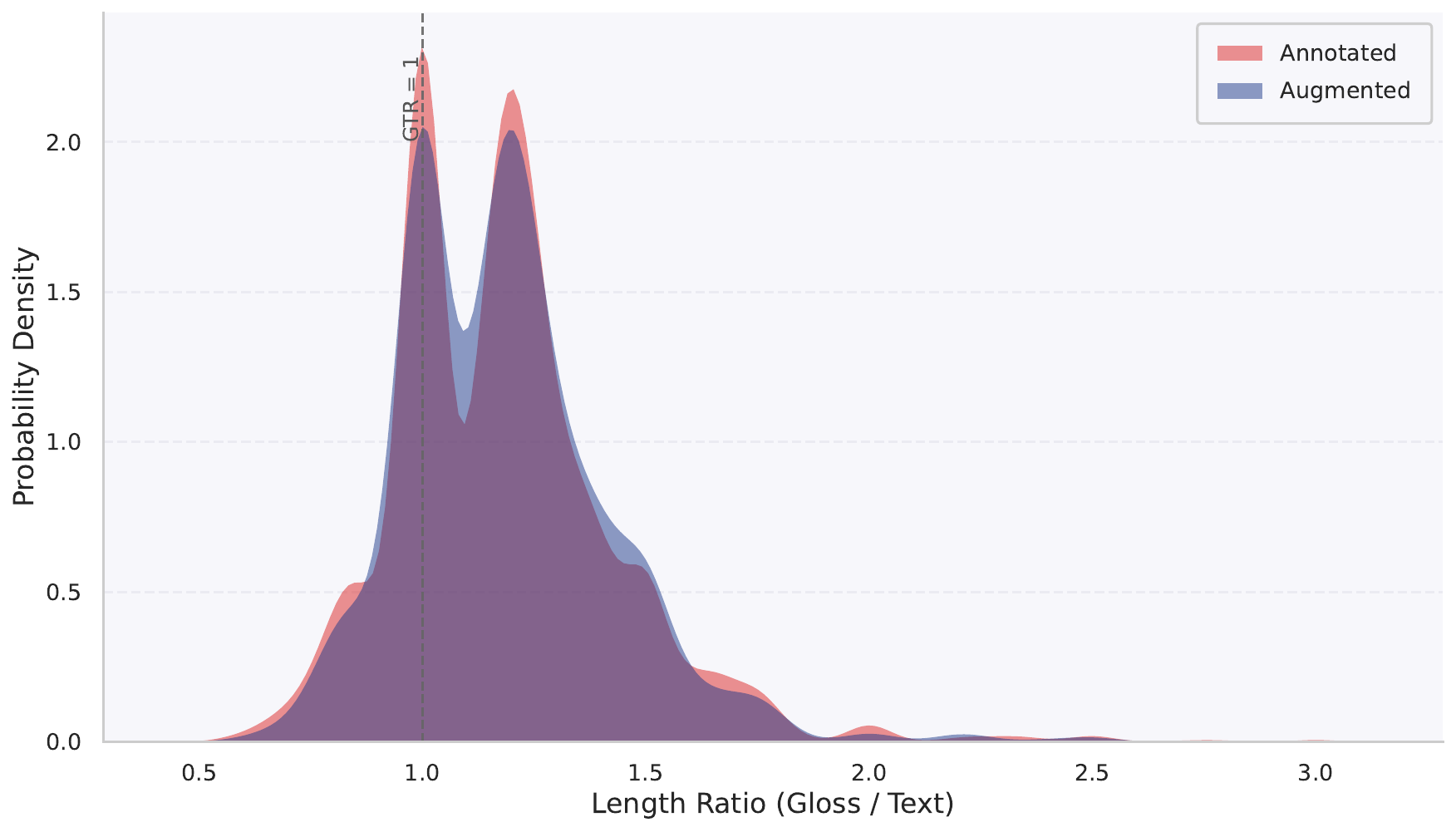}
        \caption{Gloss-to-text length ratios.}
        \label{fig:gtr_distribution}
    \end{subfigure}
    \hfill
    \begin{subfigure}[t]{0.32\textwidth}
        \centering
        \includegraphics[width=\linewidth]{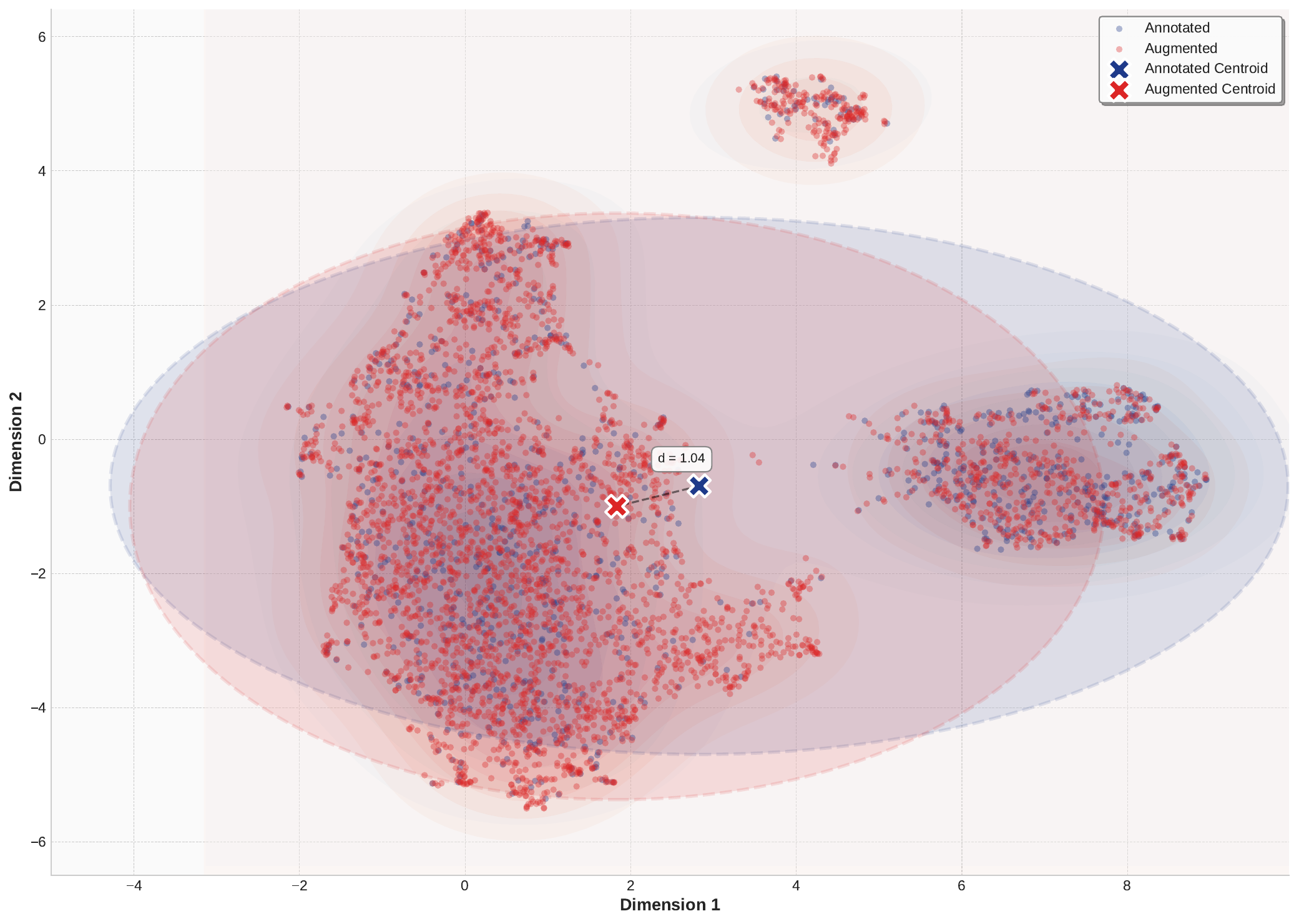}
        \caption{Semantic space projection.}
        \label{fig:umap_semantic}
    \end{subfigure}
    \hfill
    \begin{subfigure}[t]{0.32\textwidth}
        \centering
        \includegraphics[width=\linewidth]{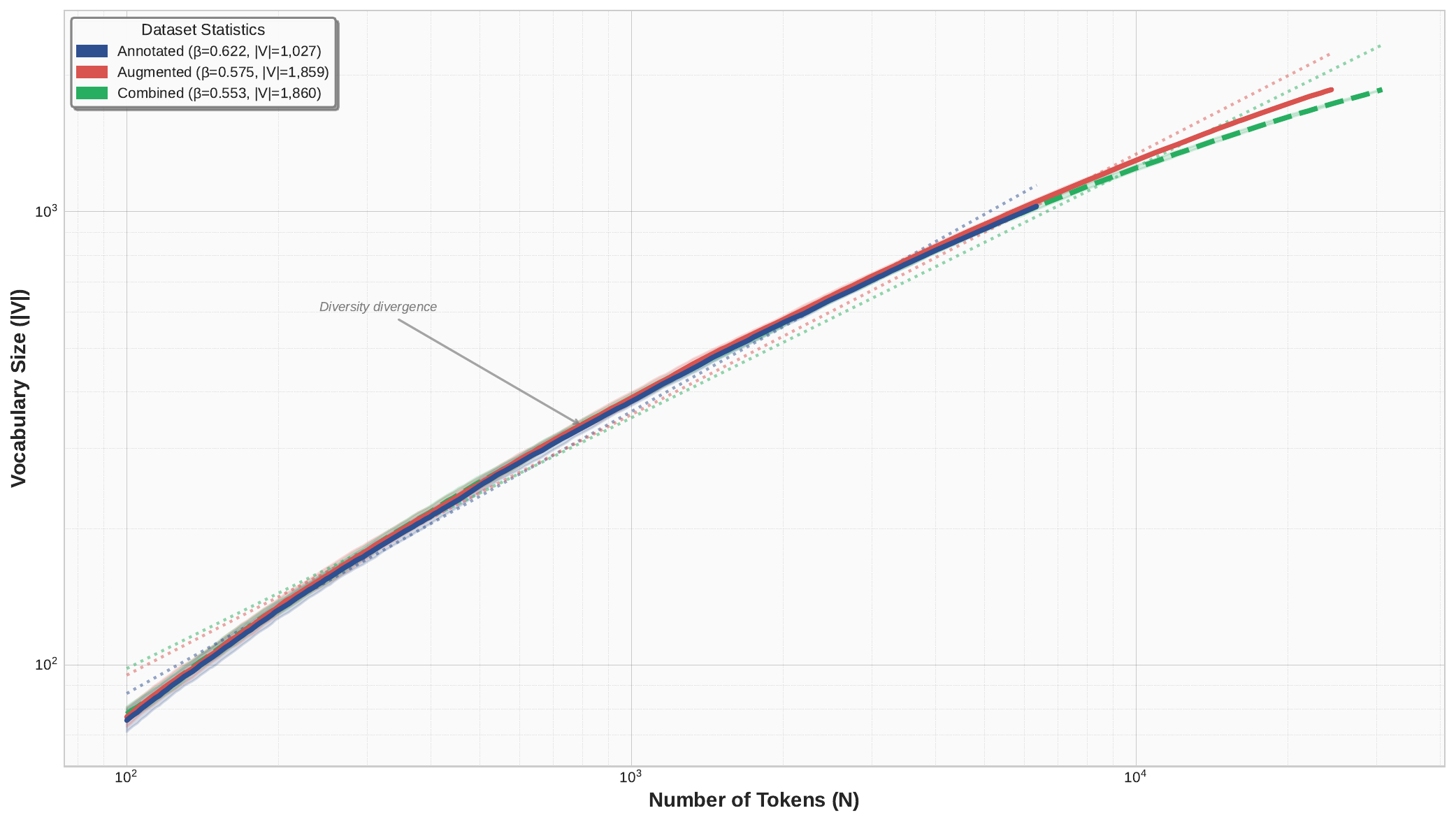}
        \caption{Lexical diversity under Heaps' Law.}
        \label{fig:heaps_law}
    \end{subfigure}

    \caption{Dataset validation using (a) sentence length distributions, (b) gloss-to-text length ratios, (c) UMAP projections of BanglaBERT embeddings, and (d) Heaps' Law vocabulary growth.}
    \label{fig:dataset_validation}
\end{figure*}

\subsubsection{Human Evaluation of Dataset}
We performed a human quality check on 300 randomly sampled gloss–sentence pairs drawn from the annotated and augmented dataset. Two BdSL signers independently rated each pair for adequacy (0–100) and fluency (1=Bad, 2=Moderate, 3=Good). The resultant statistics are in Table \ref{tab:dataset-human-eval}.
They both received professional BdSL training from Krishti, a nonprofit organization. Regarding their educational background, one evaluator was pursuing a secondary school qualification. The other evaluator had completed their higher secondary education. The evaluation was conducted in person. A trained sign language interpreter was present during the evaluation. The interpreter helped the evaluators communicate and understand the instructions. Before starting the evaluation, we explained two terms, i.e. adequacy and fluency, to the evaluators. The evaluators assigned separate scores for adequacy and fluency on their respective scales. We recorded the scores given by the evaluators. For each sample, we calculated the average separately for adequacy and fluency. No third-party adjudication or any tie-breaking method was used to resolve the disagreements between two evaluators.
The mean adequacy is 99.21 (SD = 7.11) with a median of 100. 99.3\% of samples reach the high-quality adequacy threshold ($\geq 75\%$). Fluency checks are equally positive. The overall mean fluency is 2.99 (SD = 0.11) and 98.0\% of items were rated “Good” by both evaluators. When considering both criteria, 97.7\% of pairs satisfy adequacy $\geq 75\%$ and fluency = Good. Inter-annotator agreement analysis supports the reliability of these ratings. Pearson correlation for adequacy is $r=0.728$ (p<0.001) and the mean absolute error is only 0.75 points on the 0–100 DA scale. For fluency, weighted Cohen’s $\kappa=0.495$ with exact agreement 98.7\% and adjacent agreement 100\%, indicating moderate $\kappa$. The intraclass correlation (ICC) for adequacy is 0.465, consistent with significant consistency given the large sample.

\begin{table*}[t]
\centering
\caption{Human evaluation results for the annotated + augmented dataset (300 samples). \textsuperscript{a}Adequacy $\geq 75\%$ AND Fluency = Good (both evaluators). $r$ = Pearson correlation, $\kappa_w$ = Weighted Cohen's Kappa, $\rho$ = Spearman's rho. MAE = Mean Absolute Error, Exact = Exact agreement percentage. *$p < 0.05$, **$p < 0.01$, ***$p < 0.001$.}
\label{tab:dataset-human-eval}

\resizebox{\linewidth}{!}{%
\begin{tabular}{lcccc}
\toprule
\textbf{Metric} & \textbf{Evaluator 1} & \textbf{Evaluator 2} & \textbf{Overall} & \textbf{Agreement} \\
\midrule
\multicolumn{5}{l}{\textit{\textbf{Adequacy}}} \\
Mean $\pm$ SD & 98.83 $\pm$ 8.09 & 99.58 $\pm$ 5.95 & 99.21 $\pm$ 7.11 & $r = 0.728$*** \\
Median & 100 & 100 & 100 & MAE = 0.75 \\
High Quality ($\geq 75\%$) & 298 (99.3\%) & 298 (99.3\%) & 298 (99.3\%) & --- \\
\midrule
\multicolumn{5}{l}{\textit{\textbf{Fluency}}} \\
Mean $\pm$ SD & 2.98 $\pm$ 0.14 & 2.99 $\pm$ 0.08 & 2.99 $\pm$ 0.11 & $\kappa_w = 0.495$ \\
Median & 3.0 & 3.0 & 3.0 & $\rho = 0.573$*** \\
\midrule
\multicolumn{5}{l}{\textit{\textbf{Fluency Distribution}}} \\
Bad (1) & 0 (0.0\%) & 0 (0.0\%) & 0 (0.0\%) & --- \\
Moderate (2) & 6 (2.0\%) & 2 (0.7\%) & 8 (1.3\%) & --- \\
Good (3) & 294 (98.0\%) & 298 (99.3\%) & 592 (98.7\% of 600 ratings) & Exact: 98.7\% \\
\midrule
\multicolumn{5}{l}{\textit{\textbf{Overall Quality Assessment}}} \\
Both criteria met\textsuperscript{a} & \multicolumn{2}{c}{293 (97.7\%)} & Adeq--Flu corr. & $\rho = 0.391$*** \\
\bottomrule
\end{tabular}
}
\end{table*}

\subsection{Initial Model Selection}
\label{subsec:initial_model_selection}

For the G2S and S2G synthesis tasks, we selected a compact encoder–decoder BART \citep{bart} model. The architecture was configured from scratch to be small (56M parameters) and monolingual for Bangla, rather than adopting large pre-trained checkpoints. BART (encoder–decoder, denoising) is explicitly designed for sequence-to-sequence tasks and the denoising training is particularly helpful for learning robust mappings between two modalities of text (glosses and sentences).
Although pre-trained multilingual and English-centric checkpoints (e.g., large GPT or BART variants) offer strong priors, we deliberately refrain from using them for this study because pre-trained multilingual models contain many languages and suboptimal tokenization/representations for Bangla. Training a compact, monolingual model from scratch allows us to tailor the vocabulary, subword segmentation, and inductive bias specifically to Bangla morphology and the glossing conventions used in our annotation pipeline. Large pre-trained models are memory and compute-intensive at inference, complicating deployment on resource-constrained hardware (edge devices, low-cost servers). Our target use-cases (interactive annotation tools and real-time synthesis) require small, low-latency models. Table \ref{tab:models_compare} provides a comprehensive summary of the architecture of the model selected.

\begin{table}[t]
\caption{Model configuration used for the gloss sequence to sentence modeling.}
\label{tab:models_compare}
\centering
\begin{tabular}{lcc}
\hline
\textbf{Property}  & \textbf{BART} \\
\hline
Model family  & Seq2Seq denoising LM \\
Architecture  & Encoder + decoder \\
Encoder layers  & 4 \\
Decoder layers  & 4 \\
Hidden dimension  & 512 \\
Attention heads  & 8 / 8 (enc / dec) \\
FFN dimension  & 2048 \\
Context size  & 512 \\
Vocabulary  & 52001 \\
Approx.\ parameters  & $\sim$56M \\
Training style  & Denoising / conditional generation \\
\end{tabular}
\end{table}

\subsection{Pre-training}
\label{subsec:pretraining}

Pre-training was carried out from scratch on the six tokenized corpora. To prevent large sources from dominating the pre-training we employed temperature-controlled sampling \citep{Conneau2019} across the six corpora. Let \(s_i\) denote the (byte or token) size of corpus \(i\) and \(S=\sum_j s_j\). The base sampling probability is
\begin{equation}
p_i \;=\; \frac{s_i}{S},
\end{equation}
and the temperature-adjusted probability used to draw training blocks
\begin{equation}
\tilde{p}_i \;=\; \frac{p_i^{\,1/T}}{\sum_j p_j^{\,1/T}},
\label{eq:temp_sampling}
\end{equation}
where \(T>0\) is the sampling temperature (higher \(T\) flattens the distribution and increases sampling from smaller corpora).

\subsubsection{General Pre-training Framework}
\label{subsubsec:pretraining}
Let $X = (x_1, x_2, \dots, x_T)$ denote the original Bangla token sequence sampled from the tokenized pre-training data $\mathcal{D}$, where $x_t \in \mathcal{V}$ is a token from the vocabulary. We define a stochastic corruption function $C: \mathcal{V}^T \to \mathcal{V}^{T'}$, which is a composite function of document-level and token-level noising that transforms $X$ into a noisy version $\tilde{X}$. The IsharaBART model parameterized by $\theta$, is trained as a denoising autoencoder to reconstruct the original $X$ given $\tilde{X}$. The training objective minimizes the negative log-likelihood

\begin{equation}
    \mathcal{L}(\theta) = -\mathbb{E}_{x \sim \mathcal{D}, \tilde{x} \sim C(x)} \left[ \sum_{t=1}^{|x|} \log P(x_t \mid \tilde{x}, x_{<t}; \theta) \right]
\end{equation}

Figure \ref{fig:pretraining} illustrates the general pre-training framework of the initial model with a step-by-step depiction of the full workflow.

\begin{figure*}[t]
    \centering
    \includegraphics[width=\textwidth]{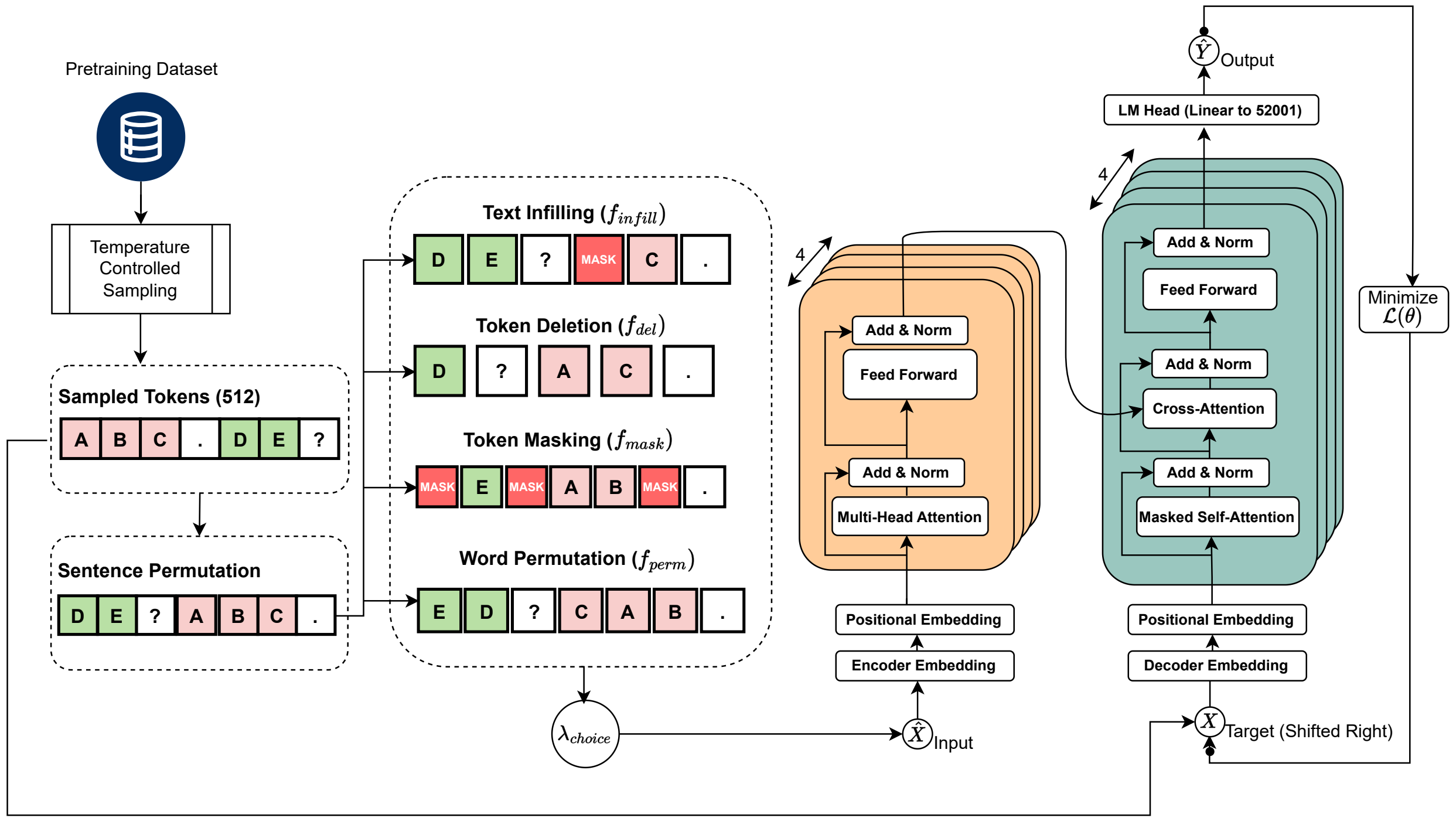}
    \caption{Proposed framework for the pre-training with denoising objectives.}
    \label{fig:pretraining}
\end{figure*}

\subsubsection{Document-Level Noising}
\label{subsubsec:docnoising}
\paragraph{Sentence permutation}
To train the model to learn logical discourse and sentence relationships, we first apply sentence permutation. This is applied with probability 1.0 \citep{bart}, shuffling sentences to disrupt global structure while preserving local coherence. Let the input sequence $X$ be segmented into $M$ sentences based on Bangla-specific delimiters (period and question mark). We define the segmented set as $\mathcal{S} = \{S_1, S_2, \dots, S_M\}$. We sample a random permutation $\sigma \in \Sigma_M$ (the set of all permutations of length $M$) with probability $p_{sent}$. The document-level corrupted sequence $\tilde{X}_{doc}$ is obtained by concatenating the permuted sentences

\begin{equation}
    \tilde{X}_{doc} = S_{\sigma(1)} \oplus S_{\sigma(2)} \oplus \dots \oplus S_{\sigma(M)}
\end{equation}
where $\oplus$ denotes sequence concatenation.

\subsubsection{Token-Level Noising}
After sentence permutation, we apply one of four distinct token-level noise transformations to $\tilde{X}_{doc}$. Let $Z = (z_1, \dots, z_L)$ be the tokens of $\tilde{X}_{doc}$. We select a transformation $f \in \{f_{infill}, f_{del}, f_{mask}, f_{perm}\}$ based on a categorical distribution $\lambda_{choice} = [0.3, 0.2, 0.2, 0.3]$.
\label{subsubsec:tokennoising}
\paragraph{Text infilling ($f_{infill}$)}
This objective teaches the model to predict spans of missing text. We sample a set of span lengths from a Poisson distribution, where the length $l \sim \text{Poisson}(\lambda=3)$. For a chosen span of tokens starting at index $i$ with length $l$, denoted as $z_{i:i+l-1}$, we replace the entire span with a single unique mask token $[MASK]$. The corrupted sequence is

\begin{equation}
   \tilde{X} = (\dots, z_{i-1}, [MASK], z_{i+l}, \dots)
\end{equation}
Unlike standard single token masking, a span of length $l$ is replaced by exactly one mask token, forcing the model to learn how many tokens are missing.

\paragraph{Token deletion ($f_{del}$)}
To improve the model's robustness against dropped inputs (common in SLT), we delete tokens without leaving a trace. Given a sequence of length $L$, we calculate the number of tokens to remove as $N = \lfloor p_{del} \cdot L \rfloor$. We uniformly sample a subset of indices $\mathcal{I}_{del} \subset \{1, \dots, L\}$ of size $N$ without replacement. The corrupted sequence is formed by retaining only the tokens whose indices are not in the deletion set

\begin{equation}
  \tilde{X} = (x_t \mid t \notin \mathcal{I}_{del})
\end{equation}
The model learns to identify positions where inputs are missing and insert the correct tokens.

\paragraph{Token masking ($f_{mask}$)}
Following the standard single token masking objective, we randomly sample a set of indices $\mathcal{M}$ with probability $p_{mask}$. Tokens at these indices are replaced with the $[MASK]$ token

\begin{equation}
\tilde{x}_t =
    \begin{cases}
    [MASK] & \text{if } t \in \mathcal{M} \\
    z_t & \text{otherwise}
    \end{cases}
\end{equation}

\paragraph{Word permutation ($f_{perm}$)}
To model the flexible word order often found in sign language gloss sequence, we introduce an additional noising technique to shuffle words within sentence boundaries. For a given sentence $S_j \subset Z$ consisting of tokens $(w_1, \dots, w_k)$, we sample a random permutation $\pi$ of $\{1, \dots, k\}$. The sentence is transformed to

\begin{equation}
    \tilde{S}_j = (w_{\pi(1)}, w_{\pi(2)}, \dots, w_{\pi(k)})
\end{equation}
Crucially, punctuation at the end of sentence is excluded from the shuffle to preserve sentence boundaries.

\subsection{Fine-tuning}
\label{subsec:finetuning}
After pre-training, we fine-tune the pre-trained weights of IsharaBART on two separate downstream tasks: G2S translation and S2G translation, enabling bidirectional translation capability. For ablation, we employ two methods of fine-tuning: full parameter fine-tuning and LoRA \citep{LoRA}. For G2S task, we fine-tune by minimizing the negative log-likelihood of the target Bangla sentence conditioned on the corresponding input gloss sequence and previously generated tokens, following standard sequence-to-sequence training for encoder-decoder language models. For the S2G task, the input is the Bangla sentence while the target becomes the corresponding gloss sequence. Given a parallel dataset $\mathcal{D}_{\text{finetune}}$ of gloss-sentence pairs $(x, y)$, the model parameters $\theta$ are optimized by minimizing the expected token-level loss
\begin{equation}
\mathcal{L}_{\text{finetune}}(\theta)
= - \mathbb{E}_{(x,y)\sim \mathcal{D}_{\text{finetune}}}
\left[
\sum_{i=1}^{|y|}
\log P\!\left(y_i \mid x, y_{<i}; \theta\right)
\right]
\end{equation}
where $x$ denotes the input token sequence, $y = (y_1,\dots,y_{|y|})$ the corresponding target token sequence, and $P(y_i \mid x, y_{<i}; \theta)$ the model’s conditional probability of the $i$-th target token given the source and all preceding target tokens.

\section{Experimental Setup}
\label{sec:experiments}

We evaluate IsharaBART on bidirectional G2S and S2G translation tasks for BdSL, assessing the impact of data augmentation and comparing against strong multilingual and monolingual baselines. Our experiments are designed to answer two key questions: (1) Does pre-training a significantly smaller BART model with denoising objectives on monolingual Bangla text improve performance on low-resource downstream task of gloss translation? (2) How does synthetic data augmentation affect model generalization in this specialized domain?

\subsection{Downstream Tasks}
We fine-tune all models on two complementary translation directions. \textbf{G2S} takes a gloss sequence representing sign language structure as input and generates the corresponding natural Bangla sentence. \textbf{S2G} is the inverse task, mapping natural Bangla text to gloss sequences.

\subsection{Baseline Models}
We compare IsharaBART against four strong baseline architectures representing diverse pre-training objectives, architectural choices. \textbf{mBART-50} \citep{mBART} is a multilingual denoising autoencoder with an encoder--decoder Transformer architecture and approximately 611M parameters in the mBART-large-50 configuration, pre-trained on 50 languages including Bangla using a BART-style denoising objective, making it a natural baseline for our monolingual BART-style pre-training while controlling for architecture and objective but varying pre-training language coverage. \textbf{mT5-Small} \citep{mT5} is a multilingual text-to-text Transformer with an encoder--decoder architecture and roughly 300M parameters, pre-trained on a massively multilingual corpus using a span-corruption objective in the T5 framework. \textbf{BanglaT5} \citep{dialogues} is a monolingual adaptation of the T5 architecture pre-trained exclusively on Bangla corpora, with approximately 247M parameters, serving as a direct monolingual baseline that contrasts our BART-style denoising pre-training with a T5-style span-corruption objective under a Bangla-only setting and at a substantially larger parameter scale than IsharaBART. \textbf{BanglaGPT} \citep{BanglaGPT} is a decoder-only autoregressive Transformer model for Bangla with 124M trainable parameters, corresponding to a GPT-2-style medium configuration adapted to Bangla, enabling us to assess how a pure generative pre-training paradigm compares to encoder-decoder architectures on gloss translation at a smaller parameter scale.

\subsection{Experimental Conditions}
IsharaBART is pre-trained on the six monolingual Bangla corpora using the denoising autoencoder objective described in Section~\ref{sec:methodology}. We conduct four fine-tuning experiments for each of the five models: G2S and S2G, each on annotated data only and on annotated + augmented data, yielding a $5 \times 2 \times 2$ design (20 fine-tuned models) that enables comprehensive analysis of data augmentation effects and task asymmetry across all five architectures. In addition, for IsharaBART alone, we separately ablate monolingual pre-training by comparing the pre-trained model against an identical architecture trained from scratch, isolating the contribution of pre-training independent of the cross-model comparison above. Detailed hyperparameters are provided in \ref{ap:training_config}.

\subsection{Evaluation Metrics}
We report recall-oriented understudy for gisting evaluation (ROUGE-1/2/L) \citep{rouge}, BLEU-1/2/3/4 \citep{bleu}, and crosslingual optimized metric for evaluation of translation (COMET) \citep{comet} scores as our primary evaluation metrics. ROUGE checks how much of the critical information from human-written text is successfully captured in the generated text. BLEU evaluates how many words or phrases in the generated text match reference text. We complement them with COMET, which evaluates the semantic meaning and contextual quality of the generated text. Model selection is performed based on BLEU-4 on the validation set, and all reported results are on the held-out test set.

\section{Results and Analysis}
\label{sec:results}

The experiments are designed to isolate three questions: whether monolingual Bangla pre-training benefits low-resource gloss--sentence translation, whether synthetic augmentation improves generalization, and whether the resulting model offers a favorable accuracy--efficiency trade-off. Across the controlled comparisons, three consistent findings emerge. First, monolingual pre-training substantially improves performance when only 1,000 annotated pairs are available and reduces the number of epochs required to reach peak validation BLEU-4. Second, augmentation improves translation quality in both directions, with a particularly large effect on S2G. Third, the 56M-parameter IsharaBART remains competitive with substantially larger baselines, achieving its strongest relative advantage on the bidirectional S2G task.

\subsection{Pre-training Ablation and Convergence}
\label{subsec:pretraining_ablation}

We first isolate the contribution of monolingual Bangla pre-training by comparing IsharaBART initialized from the proposed pre-trained checkpoint with the same architecture trained from scratch under identical downstream conditions.

The pre-training of IsharaBART, which was conducted on monolingual Bangla text, showed fast convergence. This is shown in Figure~\ref{fig:pretraining_loss}. Convergence was achieved around 5,000 steps, as evidenced by the plateau in both losses. Further iterations yielded diminishing returns (<0.1 reduction per 1,000 steps). The pre-training was conducted for a total of 22.3 GPU hours on an NVIDIA A100 (80GB) with an effective batch size of 256. Given the limitation of computational resource, going beyond 30,000 steps was deemed impractical. It would have exceeded available resources without significant gains in performance.

\begin{figure}[t]
    \centering
   \includegraphics[width=0.5\textwidth]{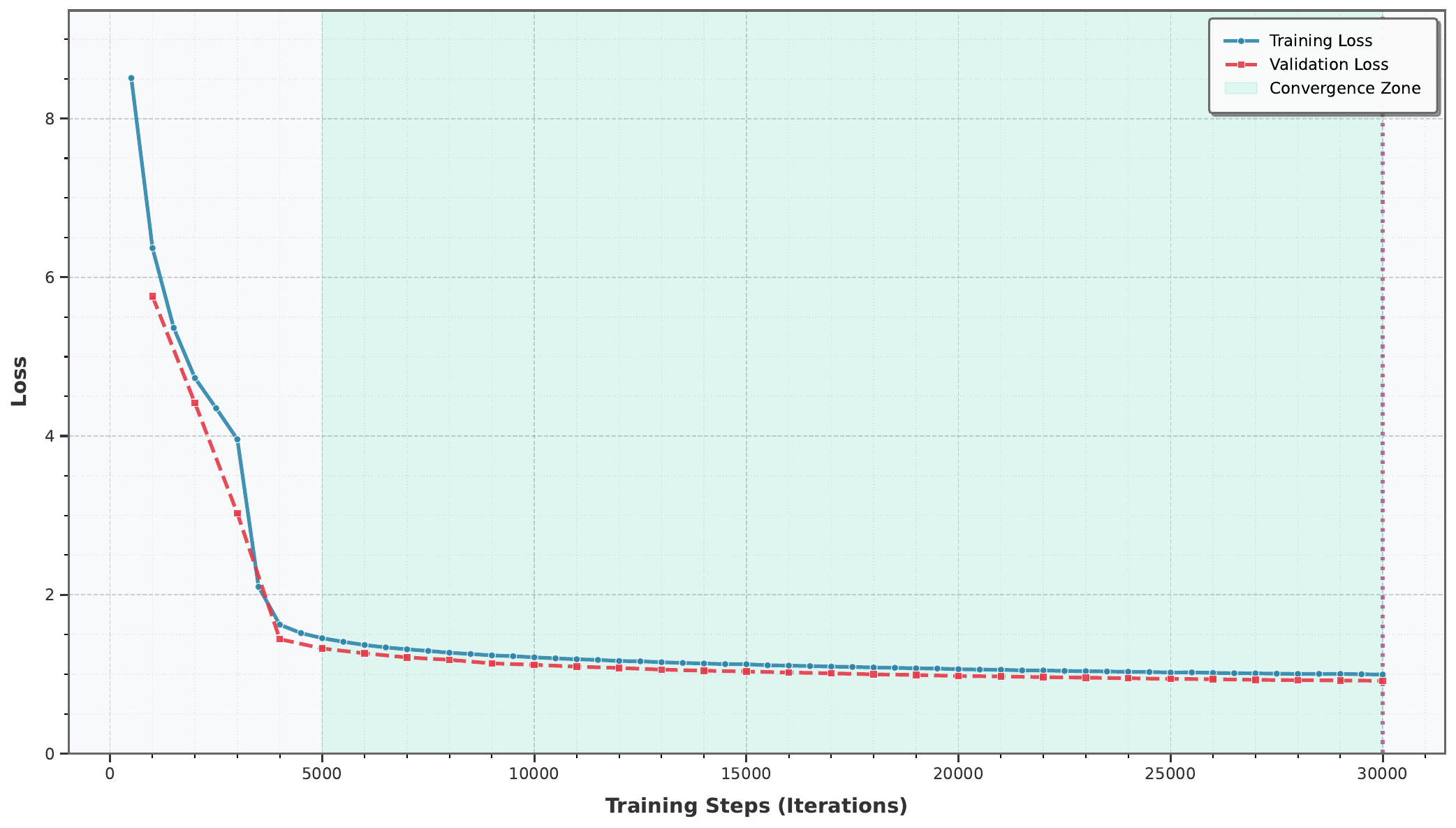}
    \caption{IsharaBART pre-training loss convergence over 30k steps. Training loss (blue) and validation loss (red) exhibit a rapid descent through 5k iterations, stabilizing with negligible improvement.}
    \label{fig:pretraining_loss}
\end{figure}

On the annotated dataset, pre-training increases G2S BLEU-4 from 19.83 to 43.30 and S2G BLEU-4 from 23.80 to 39.39, corresponding to relative improvements of 118.4\% and 65.5\%, respectively. The gains remain after augmentation: BLEU-4 increases from 45.30 to 51.74 for G2S and from 55.10 to 61.40 for S2G. Thus, pre-training is most valuable when parallel gloss-sentence data is scarce, but its benefit continues even after the training set is expanded.

\begin{figure*}[t]
    \centering
    \includegraphics[width=\textwidth]{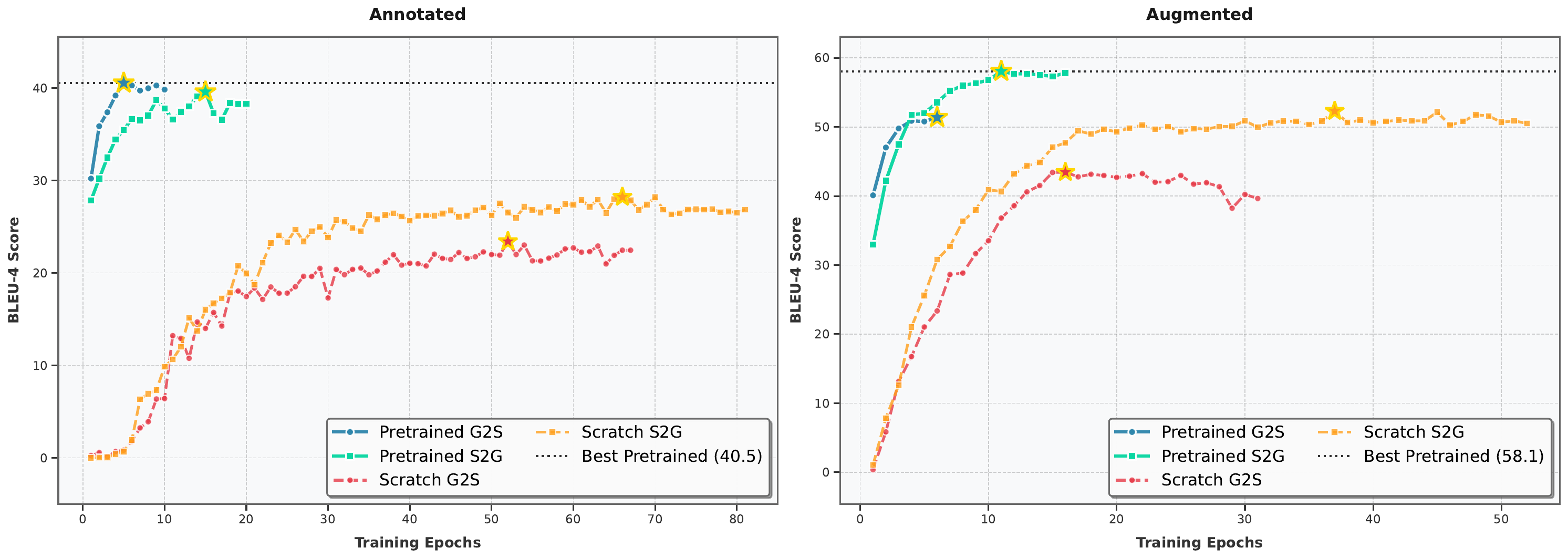}
    \caption{BLEU-4 convergence curves comparing pre-trained IsharaBART against from-scratch baselines across annotated and augmented datasets for bidirectional translation tasks. Stars indicate peak performance. The dotted horizontal line marks the best pre-trained performance for reference.}
    \label{fig:convergence_analysis}
\end{figure*}

\begin{figure*}[t]
    \centering
    \includegraphics[width=0.8\textwidth]{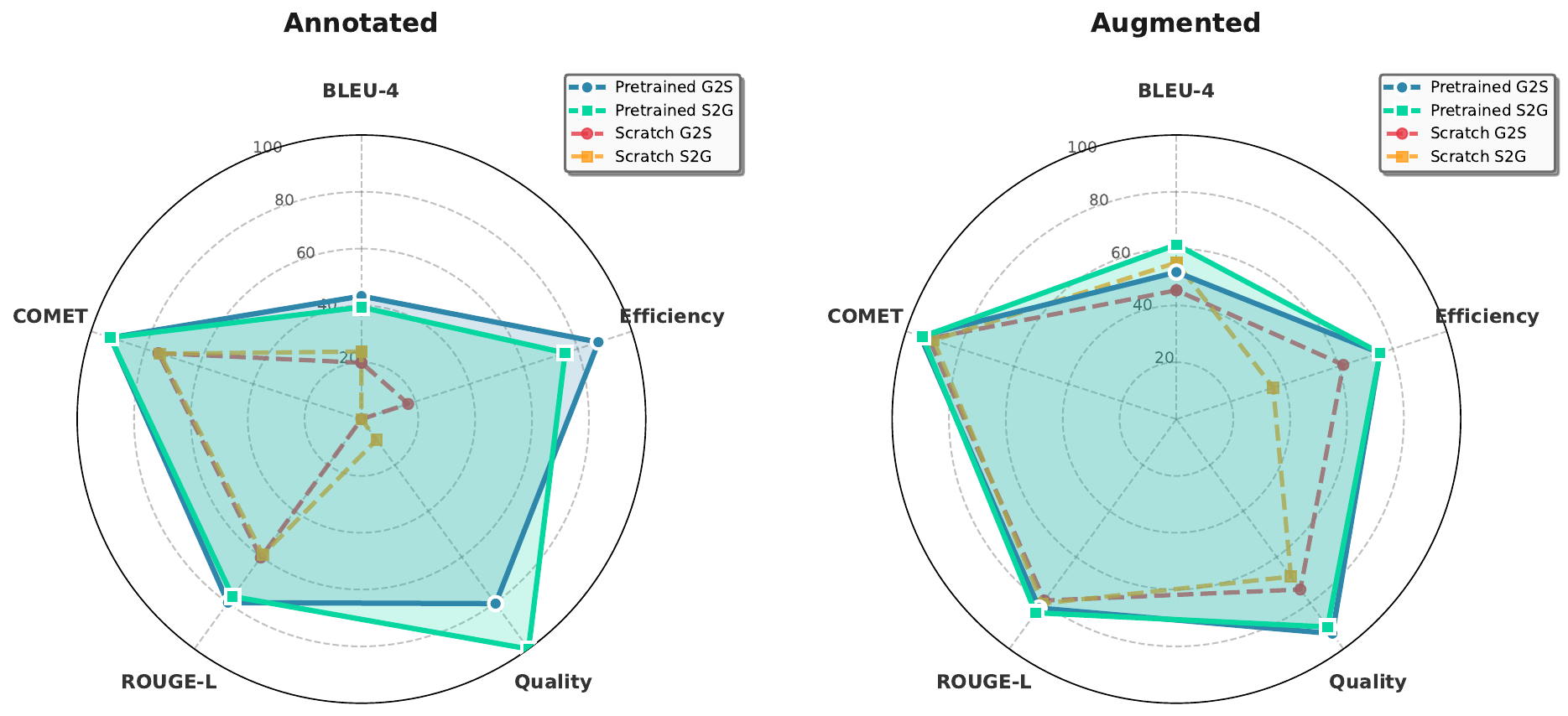}
    \caption{Multi-dimensional performance comparison between pre-trained IsharaBART and from-scratch baselines. The charts visualize five key metrics normalized to a 0--100 scale: BLEU-4, COMET, ROUGE-L, Quality (min-max normalized inverse loss), and Efficiency (percentage reduction in training epochs).}
    \label{fig:pretrain_ablation}
\end{figure*}

Pre-training also reduces the number of epochs required to reach peak validation BLEU-4 (Figure~\ref{fig:convergence_analysis}, Figure \ref{fig:pretrain_ablation}). On the annotated data, G2S reaches peak performance in 5 epochs compared with 52 for the scratch model, while S2G requires 15 rather than 66 epochs, corresponding to 90.4\% and 77.3\% fewer epochs. With augmented data, the corresponding reductions are 62.5\% for G2S (6 vs.\ 16 epochs) and 70.3\% for S2G (11 vs.\ 37 epochs). These results indicate that monolingual Bangla pre-training improves both the quality and the sample efficiency of downstream adaptation.

\subsection{G2S Translation}

On the annotated dataset, BanglaT5 achieves the highest scores (Table \ref{tab:g2s_ann_aug}). It gets 48.91 BLEU-4, 0.95 COMET, and 81.71 ROUGE-L. IsharaBART gets 43.30 BLEU-4, 0.93 COMET, and 78.32 ROUGE-L. However, IsharaBART has only 56M parameters, while BanglaT5 has 247M.

Data augmentation reduces this gap. IsharaBART reaches 51.74 BLEU-4 and 0.95 COMET after augmentation. BanglaT5 reaches 52.24 BLEU-4. The difference is only 0.50 BLEU-4 points. IsharaBART also uses 77.3\% fewer parameters. These results show that IsharaBART can achieve scores close to those of a much larger model with less computational capacity.

Augmentation raises IsharaBART's G2S BLEU-4 score from 43.30 to 51.74. This is an increase of 8.44 points. The baseline models also improve after augmentation. This suggests that the added data helps different models, not just IsharaBART. The results also show that synthetic data can help reduce the performance gap between smaller and larger models.

\begin{table*}[t]
\centering
\caption{Translation results for G2S on the annotated and augmented datasets.}
\label{tab:g2s_ann_aug}
\resizebox{\textwidth}{!}{%
\begin{tabular}{l|c|ccc|cccc|c}
\toprule
\textbf{Model} & \textbf{Params} & \textbf{R-1} & \textbf{R-2} & \textbf{R-L} &
\textbf{B-1} & \textbf{B-2} & \textbf{B-3} & \textbf{B-4} & \textbf{COMET} \\
\midrule

\multicolumn{10}{c}{\textit{\textbf{G2S -- Annotated Dataset (1,000 pairs)}}} \\
\midrule

mBART-50 & 611M & 77.86 & 63.36 & 76.63 & 65.15 & 54.64 & 46.39 & 39.93 & 0.92 \\
mT5-Small & 300M & 77.76 & 62.37 & 76.20 & 63.78 & 52.78 & 44.76 & 38.76 & 0.90 \\
BanglaT5 & 247M & \textbf{82.94} & \textbf{70.65} & \textbf{81.71} & 72.31 & \textbf{62.84} & \textbf{55.16} & \textbf{48.91} & \textbf{0.95} \\
BanglaGPT & 124M & 61.27 & 45.16 & 60.76 & 41.88 & 31.69 & 23.83 & 17.82 & 0.78 \\
\textbf{IsharaBART (ours)} & 56M & 79.86 & 65.22 & 78.32 & \textbf{72.73} & 59.29 & 50.05 & 43.30 & 0.93 \\

\midrule

\multicolumn{10}{c}{\textit{\textbf{G2S -- Augmented Dataset (4,831 pairs)}}} \\
\midrule

mBART-50 & 611M & 81.33 & 69.33 & 80.40 & 71.37 & 62.36 & 55.25 & 49.12 & 0.94 \\
mT5-Small & 300M & 83.28 & 72.06 & 82.25 & 73.62 & 64.77 & 57.56 & 51.78 & 0.94 \\
BanglaT5 & 247M & \textbf{83.74} & \textbf{72.78} & \textbf{82.60} & 74.38 & 65.40 & \textbf{58.20} & \textbf{52.24} & \textbf{0.95} \\
BanglaGPT & 124M & 72.36 & 57.72 & 71.29 & 56.87 & 46.32 & 37.97 & 31.26 & 0.88 \\
\textbf{IsharaBART (ours)} & 56M & 82.22 & 70.51 & 80.95 & \textbf{77.97} & \textbf{66.11} & 57.98 & 51.74 & \textbf{0.95} \\

\bottomrule
\end{tabular}%
}
\end{table*}

We also present qualitative examples from the test set to show the performance of IsharaBART on the G2S translation task. These examples are tabularized in Table \ref{tab:g2s-examples}.

\begin{table*}[t]
\centering
\caption{Qualitative G2S translation examples across three performance levels. English translations of input glosses are approximate and shown in uppercase.}
\label{tab:g2s-examples}

\begingroup
\fontsize{5.5}{5.9}\selectfont
\setlength{\tabcolsep}{1.5pt}
\renewcommand{\arraystretch}{0.65}
\setlength{\extrarowheight}{0pt}
\setlength{\aboverulesep}{1pt}
\setlength{\belowrulesep}{1pt}

\renewcommand{\bntrans}[2]{%
    \bn{#1}\par\vspace{1.5pt}%
    {\fontsize{4.5}{4.8}\selectfont\textit{(#2)}}%
}
\begin{tabular}{@{}
>{\raggedright\arraybackslash}p{0.28\textwidth}
>{\raggedright\arraybackslash}p{0.34\textwidth}
>{\raggedright\arraybackslash}p{0.34\textwidth}
@{}}
\toprule
\textbf{Input (Gloss)} &
\textbf{Ground Truth (Bangla)} &
\textbf{Predicted (Bangla)} \\
\midrule

\multicolumn{3}{c}{\textbf{Excellent Performance}} \\
\midrule

\bntrans{তুমি তুমি আস্তে আস্তে বড় হবে, ভালো মানুষ কাজ কাজ চেষ্টা করবে}
{YOU YOU SLOWLY SLOWLY GROW WILL, GOOD PERSON WORK WORK TRY WILL}
&
\bntrans{তুমি বড় হয়ে ভালো মানুষ হওয়ার চেষ্টা করবে।}
{You will try to become a good person when you grow up.}
&
\bntrans{তুমি বড় হয়ে ভালো মানুষ হওয়ার চেষ্টা করবে।}
{You will try to become a good person when you grow up.}
\\

\bntrans{যখন তুমি আস্তে আস্তে বড় হবে, তখন তুমি কাজ কি হবে}
{WHEN YOU SLOWLY SLOWLY GROW WILL, THEN YOU WORK WHAT WILL}
&
\bntrans{যখন তুমি বড় হবে, তখন তুমি কি করবে?}
{When you grow up, what will you do?}
&
\bntrans{যখন তুমি বড় হবে, তখন তুমি কি করবে?}
{When you grow up, what will you do?}
\\

\bntrans{শিক্ষক ছাত্র নতুন বই পড়া বলা না}
{TEACHER STUDENT NEW BOOK READ TELL NOT}
&
\bntrans{শিক্ষক ছাত্রদের নতুন বই পড়তে বলবেন না।}
{The teacher did not tell the students to read the new book.}
&
\bntrans{শিক্ষক ছাত্রদের নতুন বই পড়তে বলবেন না।}
{The teacher did not tell the students to read the new book.}
\\

\bntrans{আমার হাত একটি চিঠি আছে}
{MY HAND ONE LETTER HAVE}
&
\bntrans{আমার কাছে একটি চিঠি আছে।}
{I have a letter.}
&
\bntrans{আমার কাছে একটি চিঠি আছে।}
{I have a letter.}
\\

\bntrans{ছাত্র খেলা জন্য মাঠ যাচ্ছিল}
{STUDENT PLAY FOR FIELD GOING-PAST}
&
\bntrans{ছাত্ররা খেলার জন্য মাঠে যাচ্ছিল।}
{The students were going to the field to play.}
&
\bntrans{ছাত্ররা খেলার জন্য মাঠে যাচ্ছিল।}
{The students were going to the field to play.}
\\

\midrule
\multicolumn{3}{c}{\textbf{Moderate Performance}} \\
\midrule

\bntrans{তখন আকাশ মেঘ মেঘ ছিল না}
{THEN SKY CLOUD CLOUD WAS NOT}
&
\bntrans{তখন আকাশ মেঘলা ছিল না।}
{The sky was not cloudy at that time.}
&
\bntrans{তখন আকাশ অন্ধকার ছিল না।}
{The sky was not dark at that time.}
\\

\bntrans{তুমি তার বিষয়ে বোঝা সহজ}
{YOU HIS/HER ABOUT UNDERSTAND EASY}
&
\bntrans{আপনি ওকে ভালো করে বোঝেন।}
{You understand him/her well.}
&
\bntrans{তুমি তাকে ভালোভাবে বুঝতে।}
{You used to understand him/her well.}
\\

\bntrans{যখন সে কান্না ছিল তখন আমি তার হাত ধরা ছিল}
{WHEN HE/SHE CRY WAS, THEN I HIS/HER HAND HOLD WAS}
&
\bntrans{যখন সে কাঁদছিল, আমি তার হাত ধরেছিলাম।}
{When he/she was crying, I was holding his/her hand.}
&
\bntrans{সে যখন কাঁদছিল, তখন আমি তার হাত ধরেছিলাম।}
{When he/she was crying, then I was holding his/her hand.}
\\

\bntrans{আমি ক্লাস শেষ পরে খাওয়া হবে}
{I CLASS FINISH AFTER EAT WILL}
&
\bntrans{ক্লাসের পর খাবো।}
{Will eat after class.}
&
\bntrans{আমি ক্লাস শেষ করে খাব।}
{I will eat after finishing class.}
\\

\bntrans{নাফিস তার আগে অভ্যাস বদলানো চাওয়া ছিল}
{NAFIS HIS/HER BEFORE HABIT CHANGE WANT WAS}
&
\bntrans{নাফিস তার অভ্যাস বদলাতে চেয়েছিল।}
{Nafis wanted to change his habit.}
&
\bntrans{নাফিস তার অভ্যাস পরিবর্তন করতে চেয়েছিল।}
{Nafis wanted to change his habit.}
\\

\midrule
\multicolumn{3}{c}{\textbf{Poor Performance (Limitations)}} \\
\midrule

\bntrans{প্রথমে আমি যাওয়া পরে পড়া বসা}
{FIRST I GO AFTER STUDY SIT}
&
\bntrans{প্রথমে আমি গেলাম, তারপর পড়তে বসলাম।}
{First, I went, then sat down to study.}
&
\bntrans{প্রথমে আমি পড়ি, তারপর পড়তে বসলাম।}
{First, I studied, then sat down to study.}
\\

\bntrans{আমি সংবাদ পত্র পড়ি}
{I NEWS PAPER READ}
&
\bntrans{আমি খবরের কাগজ পড়ি।}
{I read the newspaper.}
&
\bntrans{আমি পত্র পড়ি।}
{I read a letter.}
\\

\bntrans{আমি যাকে বন্ধু চিন্তা ছিল সে একজন প্রতারক}
{I WHO FRIEND THINK WAS, HE ONE FRAUD}
&
\bntrans{বন্ধু যাকে ভাবছি সে আসলে একজন প্রতারক।}
{The friend I thought I knew was actually a fraud.}
&
\bntrans{যে বন্ধুকে আমি প্রতারক মনে করি।}
{The friend whom I consider a fraud.}
\\

\bntrans{তুমি কম্পাস ওখানে একা চলা হবে}
{YOU CAMPUS THERE ALONE WALK WILL}
&
\bntrans{আপনি একা হেঁটে ক্যাম্পাসে যাবেন।}
{You will go to the campus alone on foot.}
&
\bntrans{তুমি কম্পাস ওখানে চলবে একলা।}
{You will go to campus alone.}
\\

\bntrans{তুমি বৃষ্টি থামা পরে কাজ শুরু করো}
{YOU RAIN STOP AFTER WORK START DO}
&
\bntrans{তুমি বৃষ্টি থামলে কাজ শুরু করে দাও।}
{You will start working after the rain stops.}
&
\bntrans{আপনি বৃষ্টি থামিয়ে কাজ শুরু করুন।}
{You will start working after stopping the rain.}
\\

\bottomrule
\end{tabular}
\endgroup
\end{table*}

\subsection{S2G Translation}
The S2G results show a different pattern (Table \ref{tab:s2g_ann_aug}). On the annotated dataset, IsharaBART achieves the highest BLEU-4 (39.39) and COMET (0.94). It closely surpasses BanglaT5 at 39.29 BLEU-4 and 0.93 COMET. After augmentation, IsharaBART reaches 61.40 BLEU-4, the highest score among all five models, while mBART-50 reaches 60.82 and BanglaT5 51.57. IsharaBART therefore outperforms the 611M-parameter mBART-50 by 0.58 BLEU-4 while using approximately 91\% fewer parameters.

The effect of augmentation is particularly strong for S2G. IsharaBART improves from 39.39 to 61.40 BLEU-4, a gain of 22.01 points. This is considerably larger than the corresponding G2S gain of 8.44 points. This suggests that additional lexical and structural variation is especially valuable when learning the more specialized mapping from natural Bangla sentences to gloss sequences. The difference between the two directions also indicates that S2G remains the more difficult task for parameter-efficient fine-tuning. We also present qualitative examples from the test set that are tabularized in Table \ref{tab:s2g-examples}.

\begin{table*}[t]
\centering
\caption{Translation results for S2G on the annotated and augmented datasets.}
\label{tab:s2g_ann_aug}
\resizebox{\textwidth}{!}{%
\begin{tabular}{l|c|ccc|cccc|c}
\toprule
\textbf{Model} & \textbf{Params} & \textbf{R-1} & \textbf{R-2} & \textbf{R-L} &
\textbf{B-1} & \textbf{B-2} & \textbf{B-3} & \textbf{B-4} & \textbf{COMET} \\
\midrule

\multicolumn{10}{c}{\textit{\textbf{S2G -- Annotated Dataset (1,000 pairs)}}} \\
\midrule

mBART-50 & 611M & 78.84 & 63.35 & 76.74 & 70.55 & 56.89 & 45.96 & 37.03 & 0.92 \\
mT5-Small & 300M & 74.88 & 60.78 & 73.60 & 65.35 & 52.81 & 43.23 & 35.96 & 0.87 \\
BanglaT5 & 247M & \textbf{80.06} & \textbf{65.41} & \textbf{78.54} & \textbf{73.60} & \textbf{60.20} & \textbf{49.09} & 39.29 & 0.93 \\
BanglaGPT & 124M & 58.60 & 43.38 & 58.01 & 45.97 & 33.23 & 23.54 & 16.19 & 0.77 \\
\textbf{IsharaBART (ours)} & 56M & 77.16 & 63.01 & 75.78 & 69.08 & 56.71 & 47.18 & \textbf{39.39} & \textbf{0.94} \\

\midrule

\multicolumn{10}{c}{\textit{\textbf{S2G -- Augmented Dataset (4,831 pairs)}}} \\
\midrule

mBART-50 & 611M & \textbf{84.65} & \textbf{76.10} & \textbf{83.78} & 80.91 & 72.60 & 66.16 & 60.82 & 0.94 \\
mT5-Small & 300M & 83.70 & 74.85 & 82.78 & 79.67 & 71.19 & 64.42 & 58.92 & 0.93 \\
BanglaT5 & 247M & 82.46 & 71.65 & 81.01 & 77.35 & 66.82 & 58.42 & 51.57 & 0.94 \\
BanglaGPT & 124M & 71.87 & 58.75 & 70.54 & 63.20 & 51.25 & 42.75 & 35.96 & 0.88 \\
\textbf{IsharaBART (ours)} & 56M & 84.24 & 75.67 & 83.07 & \textbf{81.06} & \textbf{72.93} & \textbf{66.55} & \textbf{61.40} & \textbf{0.94} \\

\bottomrule
\end{tabular}%
}
\end{table*}

\begin{table*}[t]
\centering
\caption{Qualitative S2G translation examples across three performance levels. English translations of input sentences are shown below the Bangla, and gloss translations are approximate uppercase renderings.}
\label{tab:s2g-examples}

\begingroup
\fontsize{5.5}{5.9}\selectfont
\setlength{\tabcolsep}{1.5pt}
\renewcommand{\arraystretch}{0.65}
\setlength{\extrarowheight}{0pt}
\setlength{\aboverulesep}{1pt}
\setlength{\belowrulesep}{1pt}

\renewcommand{\bntrans}[2]{%
    \bn{#1}\par\vspace{1.5pt}%
    {\fontsize{4.5}{4.8}\selectfont\textit{(#2)}}%
}

\begin{tabular}{@{}
>{\raggedright\arraybackslash}p{0.28\textwidth}
>{\raggedright\arraybackslash}p{0.34\textwidth}
>{\raggedright\arraybackslash}p{0.34\textwidth}
@{}}
\toprule
\textbf{Input (Bangla)} &
\textbf{Ground Truth (Gloss)} &
\textbf{Predicted (Gloss)} \\
\midrule

\multicolumn{3}{c}{\textbf{Excellent Performance}} \\
\midrule

\bntrans{সে কাজে হাত দেয়ার আগেই কাজ শেষহয়ে গিয়েছিল।}
{He/she had finished the work before even starting it.}
&
\bntrans{সে কাজ হাত শুরু আগেই কাজ শেষ ছিল}
{HE/SHE WORK HAND START BEFORE WORK FINISHED WAS}
&
\bntrans{সে কাজ হাত শুরু আগেই কাজ শেষ ছিল}
{HE/SHE WORK HAND START BEFORE WORK FINISHED WAS}
\\

\bntrans{শিক্ষক ছাত্রদের নতুন বই পড়তে বলবেন না।}
{The teacher will not ask the students to read the new book.}
&
\bntrans{শিক্ষক ছাত্র নতুন বই পড়া বলা না}
{TEACHER STUDENT NEW BOOK READ TELL NOT}
&
\bntrans{শিক্ষক ছাত্র নতুন বই পড়া হবে না}
{TEACHER STUDENT NEW BOOK READ WILL NOT}
\\

\bntrans{ডাক্তার কোনো নতুন ওষুধ দেবেন না।}
{The doctor will not prescribe any new medicine.}
&
\bntrans{ডাক্তার কোন নতুন ওষুধ লিখে দেওয়া না}
{DOCTOR ANY NEW MEDICINE PRESCRIBE GIVE NOT}
&
\bntrans{ডাক্তার কোন নতুন ওষুধ লিখে দেওয়া না}
{DOCTOR ANY NEW MEDICINE PRESCRIBE GIVE NOT}
\\

\bntrans{নাফিস তার অভ্যাস বদলাতে চেয়েছিল।}
{Nafis wanted to change his habit.}
&
\bntrans{নাফিস তার আগে অভ্যাস বদলানো চাওয়া ছিল}
{NAFIS HIS/HER BEFORE HABIT CHANGE WANT WAS}
&
\bntrans{নাফিস তার আগে অভ্যাস বদলানো চাওয়া ছিল}
{NAFIS HIS/HER BEFORE HABIT CHANGE WANT WAS}
\\

\bntrans{যখন সে কাঁদছিল, আমি তার হাত ধরেছিলাম।}
{When he/she was crying, I was holding his/her hand.}
&
\bntrans{যখন সে কান্না ছিল তখন আমি তার হাত ধরা ছিল}
{WHEN HE/SHE CRY WAS THEN I HIS/HER HAND HOLD WAS}
&
\bntrans{যখন সে কান্না ছিল তখন আমি তার হাত ধরা ছিল}
{WHEN HE/SHE CRY WAS THEN I HIS/HER HAND HOLD WAS}
\\

\midrule
\multicolumn{3}{c}{\textbf{Moderate Performance}} \\
\midrule

\bntrans{যখন তুমি বড় হবে, তখন তুমি কি করবে?}
{When you grow up, what will you do?}
&
\bntrans{যখন তুমি আস্তে আস্তে বড় হবে, তখন তুমি কাজ কি হবে}
{WHEN YOU SLOWLY SLOWLY GROW WILL THEN YOU WORK WHAT WILL}
&
\bntrans{যখন তুমি আস্তে আস্তে বড় হবে, তখন তুমি কি কাজ হবে}
{WHEN YOU SLOWLY SLOWLY GROW WILL THEN YOU WHAT WORK WILL}
\\

\bntrans{আমাকে ঐ রেস্তোরাঁয় খেতে মানা করেছে।}
{I was told not to eat at that restaurant.}
&
\bntrans{আমাকে ওই রেস্তোরাঁয় খাওয়া করা নিষেধ বলা হয়েছে}
{TO ME THAT RESTAURANT EAT DO FORBIDDEN SAY HAS BEEN}
&
\bntrans{আমাকে ওই রেস্তোরাঁয় খাওয়া নিষেধ বলা হয়েছে}
{TO ME THAT RESTAURANT EAT FORBIDDEN SAY HAS BEEN}
\\

\bntrans{তুমি আমাকে ছেড়ে চলে যেতে পারবে না।}
{You cannot leave me.}
&
\bntrans{তুমি আমাকে বাদ যাওয়া পারবে না}
{YOU ME LEAVE GO CAN NOT}
&
\bntrans{তুমি আমাকে বাদ চলে যাওয়া পারবে না}
{YOU ME LEAVE GO GO CAN NOT}
\\

\bntrans{আমি বুঝতে পারলাম না আপনি কি বলতে চাইছেন।}
{I could not understand what you wanted to say.}
&
\bntrans{আমি বুঝা নাই আপনি কি বলা চাওয়া ছিল}
{I UNDERSTAND NOT YOU WHAT SAY WANT WAS}
&
\bntrans{আমি বুঝা পারা না আপনি কি বলা চাওয়া আছে}
{I UNDERSTAND CAN NOT YOU WHAT SAY WANT IS}
\\

\bntrans{আপনি আসার সময়ে আমি ডিনার প্রস্তুত করে রেখেছিলাম।}
{I had prepared dinner by the time you arrived.}
&
\bntrans{আপনি আসার আগে আমি খাবার তৈরি করে রাখা ছিল}
{YOU ARRIVE BEFORE I FOOD PREPARE KEEP WAS}
&
\bntrans{আপনি আসার আগে আমি রাখ খাবার তৈরি করে রেখেছিলাম}
{YOU ARRIVE BEFORE I KEEP FOOD PREPARE KEEP WAS}
\\

\midrule
\multicolumn{3}{c}{\textbf{Poor Performance (Limitations)}} \\
\midrule

\bntrans{আমরা ছোটবেলায় পুলিশ সাজতাম।}
{We used to dress up as police officers when we were kids.}
&
\bntrans{আমরা ছোট সময় পুলিশ সাজা ছিল}
{WE CHILDHOOD POLICE DRESS UP USED TO}
&
\bntrans{আমরা পুলিশ অফিসার ছিল}
{WE POLICE OFFICER WAS}
\\

\bntrans{ক্লাসের পর খাবো।}
{I will eat after class.}
&
\bntrans{আমি ক্লাস শেষ পরে খাওয়া হবে}
{I CLASS FINISH AFTER EAT WILL}
&
\bntrans{ক্লাস ভিতর কি হবে}
{CLASS INSIDE WHAT WILL}
\\

\bntrans{তুমি বৃষ্টি থামলে কাজ শুরু করে দাও।}
{You will start working after the rain stops.}
&
\bntrans{তুমি বৃষ্টি থামা পরে কাজ শুরু করো}
{YOU RAIN STOP AFTER WORK START DO}
&
\bntrans{তুমি বৃষ্টি থামা পরে তাকে শুরু করো}
{YOU RAIN STOP AFTER THAT START DO}
\\

\bntrans{শিক্ষার্থীদের বই পড়ুন নিয়মিত।}
{Read the students' books regularly.}
&
\bntrans{শিক্ষার্থী বই পড় নিয়মিত নিয়মিত}
{STUDENT BOOK READ REGULARLY REGULARLY}
&
\bntrans{শিক্ষার্থী ওরা বই পড়া প্রতিদিন}
{STUDENT THEY BOOK READ EVERY DAY}
\\

\bntrans{আমাকে মা জামাটা কিনে দিতো, যদি আমি চাইতাম।}
{My mother would buy me the dress if I wanted it.}
&
\bntrans{যদি আমি জামা কেনা চাই, মা আমাকে জামা কিনে দেওয়া হবে}
{IF I DRESS BUY WANT MOTHER ME DRESS BUY GIVE WILL}
&
\bntrans{যদি আমি জামা চাই মা আমাকে জামা কিনে দেওয়া হবে}
{IF I DRESS WANT MOTHER ME DRESS BUY GIVE WILL}
\\

\bottomrule
\end{tabular}
\endgroup
\end{table*}
IsharaBART demonstrates that competitive or superior performance is achievable with far fewer parameters (56M) than standard multilingual models (300-611M) when pre-training is carefully targeted to the task language. However, the task-dependent behavior suggests that no single architecture or pre-training strategy uniformly excels across all aspects of bidirectional gloss translation, and that the choice of model should be guided by the specific deployment scenario.

\subsection{Human Evaluation}
We conducted a human evaluation of IsharaBART in both G2S and S2G translation settings by the same two BdSL signers described in Section.  They independently assessed 50 randomly sampled outputs for adequacy and fluency using a previously defined five point rating scale.A professionally trained sign language interpreter assisted us to facilitate communication. We conducted the evaluation in person and blind to the identity of the model. We calculated the mean of the two ratings for each criterion and evaluated sample. The results are summarized in Table \ref{tab:human-evaluation}. The results therefore reflect the average of the two evaluators' individual assessments.

Overall, IsharaBART demonstrates strong semantic quality across both directions. The system achieves a high mean adequacy score of 88.8, with a median of 100 and 84\% of samples rated as high quality ($\geq 75\%$) (Table \ref{tab:human-evaluation}). Similarly, S2G translation achieves a mean adequacy of 87.8, with 76\% of outputs meeting the high-quality threshold. These results indicate that the model consistently preserves the core meaning of BdSL inputs.

In both directions, the median fluency is 3 (Good), with 77\% of G2S and 73\% of S2G outputs rated as fluent. The slightly lower fluency in S2G translation is expected. Gloss generation is more constrained and sensitive to ordering and deletion. Importantly, adequacy–fluency correlations are strong and statistically significant in both settings ($\rho=0.694$ for G2S and $\rho=0.636$ for S2G). This suggests that semantically accurate translations are also perceived as more fluent by human evaluators. Inter-annotator agreement ranges from moderate to substantial, with Pearson correlations of 0.574 (G2S) and 0.450 (S2G) for adequacy, and weighted kappa values of 0.364 and 0.443 for fluency. IsharaBART satisfies both criteria for 66\% of G2S outputs and 56\% of S2G outputs when jointly considering adequacy and fluency.

\begin{table*}[t]
\centering
\scriptsize
\setlength{\tabcolsep}{3pt}
\renewcommand{\arraystretch}{0.85}
\setlength{\aboverulesep}{3pt}
\setlength{\belowrulesep}{3pt}
\caption{Human evaluation of IsharaBART for G2S and S2G translation.Two BdSL signers evaluated 50 randomly sampled translations 
on adequacy (0--100 scale) and fluency (1=Bad, 2=Moderate, 3=Good).  \textsuperscript{a}Adequacy $\geq 75\%$ AND Fluency = Good (both evaluators). $r$ = Pearson correlation, $\kappa_w$ = Weighted Cohen's Kappa, $\rho$ = Spearman's rho. MAE = Mean Absolute Error, Exact = Exact agreement percentage. *$p < 0.05$, **$p < 0.01$, ***$p < 0.001$.}
\label{tab:human-evaluation}
\resizebox{\textwidth}{!}{%
\begin{tabular}{l|cccc|cccc}
\toprule
& \multicolumn{4}{c|}{\textbf{G2S}} &
  \multicolumn{4}{c}{\textbf{S2G}} \\
\textbf{Metric} &
\textbf{Eval. 1} & \textbf{Eval. 2} & \textbf{Overall} & \textbf{Agreement} &
\textbf{Eval. 1} & \textbf{Eval. 2} & \textbf{Overall} & \textbf{Agreement} \\
\midrule

\textit{Adequacy} \\
Mean $\pm$ SD &
$85.0\pm30.3$ & $92.5\pm22.7$ & $88.8\pm26.9$ & $r=0.574^{***}$ &
$83.5\pm30.1$ & $92.0\pm21.1$ & $87.8\pm26.2$ & $r=0.450^{***}$ \\

Median &
100 & 100 & 100 & MAE = 11.5 &
100 & 100 & 100 & MAE = 12.5 \\

High quality ($\geq75\%$) &
42 (84\%) & 46 (92\%) & 42 (84\%) & --- &
38 (76\%) & 38 (76\%) & 38 (76\%) & --- \\

\midrule
\textit{Fluency} \\

Mean $\pm$ SD &
$2.64\pm0.66$ & $2.78\pm0.46$ & $2.71\pm0.57$ & $\kappa_w=0.364^{*}$ &
$2.66\pm0.69$ & $2.56\pm0.70$ & $2.61\pm0.69$ & $\kappa_w=0.443$ \\

Median &
3.0 & 3.0 & 3.0 & $\rho=0.358^{*}$ &
3.0 & 3.0 & 3.0 & $\rho=0.460^{***}$ \\

\midrule
\textit{Fluency distribution} \\

Bad (1) &
5 (10\%) & 1 (2\%) & 6 (6\%) & --- &
6 (12\%) & 6 (12\%) & 12 (12\%) & --- \\

Moderate (2) &
8 (16\%) & 9 (18\%) & 17 (17\%) & --- &
5 (10\%) & 10 (20\%) & 15 (15\%) & --- \\

Good (3) &
37 (74\%) & 40 (80\%) & 77 (77\%) & Exact: 78\% &
39 (78\%) & 34 (68\%) & 73 (73\%) & Exact: 70\% \\

\midrule
\textit{Overall quality} \\

Both criteria met\textsuperscript{a} &
--- & --- & 33 (66\%) & $\rho=0.694^{***}$ &
--- & --- & 28 (56\%) & $\rho=0.636^{***}$ \\

\bottomrule
\end{tabular}%

}
\end{table*}

\subsection{Fine-tuning Ablation Study}
\label{subsec:lora_ablation}

To evaluate parameter-efficient fine-tuning for SLT, we compared full fine-tuning against LoRA across all experimental configurations. Table~\ref{tab:lora_comparison} presents comprehensive results.

\begin{table*}[t]
\centering
\scriptsize
\setlength{\tabcolsep}{5pt}
\renewcommand{\arraystretch}{1.15}
\caption{Comparison of full fine-tuning and low-rank adaptation (LoRA) for IsharaBART across annotated and augmented data. BLEU-4 and ROUGE-L are reported on a 0--100 scale, while COMET is reported on its native 0--1 scale. Changes are reported relative to full fine-tuning as absolute and relative differences; negative values indicate degradation under LoRA.}
\label{tab:lora_comparison}
\resizebox{\textwidth}{!}{
\begin{tabular}{ll l c c c}
\toprule
\textbf{Dataset} &
\textbf{Translation Direction} &
\textbf{Fine-Tuning Strategy} &
\textbf{BLEU-4} &
\textbf{COMET} &
\textbf{ROUGE-L} \\
\midrule

\multirow{3}{*}{Annotated}
& \multirow{3}{*}{G2S}
& Full fine-tuning
& 43.30
& 0.9279
& 78.32 \\

& & LoRA
& 31.17
& 0.8973
& 72.27 \\

& & \textit{Change}
& \textit{$-12.13$ ($-28.01\%$)}
& \textit{$-0.0306$ ($-3.30\%$)}
& \textit{$-6.05$ ($-7.72\%$)} \\

\midrule

\multirow{3}{*}{Annotated}
& \multirow{3}{*}{S2G}
& Full fine-tuning
& 39.39
& 0.9283
& 75.78 \\

& & LoRA
& 22.15
& 0.9221
& 68.97 \\

& & \textit{Change}
& \textit{$-17.24$ ($-43.77\%$)}
& \textit{$-0.0062$ ($-0.67\%$)}
& \textit{$-6.81$ ($-8.99\%$)} \\

\midrule

\multirow{3}{*}{Augmented}
& \multirow{3}{*}{G2S}
& Full fine-tuning
& 51.74
& 0.9471
& 80.95 \\

& & LoRA
& 36.31
& 0.9199
& 73.06 \\

& & \textit{Change}
& \textit{$-15.43$ ($-29.82\%$)}
& \textit{$-0.0272$ ($-2.87\%$)}
& \textit{$-7.89$ ($-9.75\%$)} \\

\midrule

\multirow{3}{*}{Augmented}
& \multirow{3}{*}{S2G}
& Full fine-tuning
& 61.40
& 0.9388
& 83.07 \\

& & LoRA
& 29.37
& 0.9185
& 69.55 \\

& & \textit{Change}
& \textit{$-32.03$ ($-52.17\%$)}
& \textit{$-0.0203$ ($-2.16\%$)}
& \textit{$-13.52$ ($-16.27\%$)} \\

\bottomrule
\end{tabular}
}
\end{table*}

LoRA provides a significant parameter-efficiency advantage for IsharaBART. LoRA reduces the number of trainable parameters to approximately 0.7\% of the full model. However, this reduction in trainable parameters comes with a consistent decrease in translation performance across all evaluated settings. As shown in Table~\ref{tab:lora_comparison}, BLEU-4 decreases by 28.0--52.2\% relative to full fine-tuning. The degradation is particularly significant for S2G, where on annotated data performance decreases from 39.39 to 22.15 BLEU-4 (-43.8\%). On the augmented setting BLEU-4 decreases from 61.40 to 29.37 (-52.2\%). In contrast, the corresponding G2S reductions are smaller, at 28.0\% and 29.8\%, respectively. Thus, the results indicate a clear accuracy--parameter-efficiency trade-off, with the cost of parameter-efficient adaptation depending strongly on the translation direction.

This behavior is consistent with recent empirical evidence that LoRA does not necessarily recover the performance of full fine-tuning when the downstream task requires substantial parameter adaptation. Under standard low-rank configurations, LoRA substantially underperforms full fine-tuning in several adaptation settings, and the parameter updates learned by full fine-tuning have ranks 10--100 times larger than typical LoRA configurations~\citep{Biderman2024}. LoRA and full fine-tuning produce structurally different modifications to pretrained weights~\citep{Shuttleworth2025}. Vanilla LoRA can retain substantially less task-specific translation performance than full fine-tuning. This indicates that the LoRA constraint may prevent the model from making sufficiently expressive parameter updates~\citep{Su2024, Zeng2024}.

The difference in performance for different translation direction provides an additional insight. LoRA performs considerably better for G2S than for S2G. This suggests that the required task-specific adaptation may differ between the two directions. A plausible explanation is that G2S maps gloss sequences to natural Bangla sentences, an output space that is closer to the natural-language used during pre-training. In contrast, S2G maps natural Bangla sentences into gloss sequences, which has a more specialized and structurally difficult representation.

\subsection{Dataset Comparison}
To compare our dataset within the broader landscape of SLT resources, we compare it with prominent benchmarks and recent BdSL datasets. Table~\ref{tab:dataset-comparison} summarizes key attributes such as size, domain, and SOTA performance metrics for G2S and S2G tasks where available. Our dataset has a modest scale of approximately 4,800 pairs (including augmentation). It achieves superior BLEU-4 scores (51.7 for G2S and 61.4 for S2G in augmented data) compared to larger benchmarks. This highlights the efficiency of our approach in low-resource settings, particularly for under-resourced languages like BdSL. In terms of BdSL, datasets like IsharaKhobor and Bangla-SGP provide valuable but domain-limited parallels.

\begin{table*}[t]
\centering
\caption{Overview of major gloss-sentence datasets and associated SOTA BLEU-4 performance for G2S and S2G tasks. IsharaBART achieves substantially higher scores than prior Bangla BdSL resources despite comparable scale, highlighting the effectiveness of Bangla-specific pre-training and augmentation.}
\label{tab:dataset-comparison}
\resizebox{\textwidth}{!}{%
\begin{tabular}{llrlcc}
\toprule
\textbf{Dataset} & \textbf{Language} & \textbf{Size} & \textbf{Domain} & \textbf{G2S} & \textbf{S2G} \\
\midrule
PHOENIX-14T \citep{rwthG2S,rwthS2G} & German (DGS) & 8,257 & Weather & 18.75 & 20.68 \\
How2Sign \citep{How2Sign} & English (ASL) & 35,000 & Instructional & 8.0 & --- \\
iSign \citep{iSign} & English (ISL) & 118,000 & Open & 17.6 & --- \\
IsharaKhobor \citep{isharaKhobor} & Bangla (BdSL) & 5,642 & News & --- & --- \\
Bangla-SGP \citep{BanglaSGP} & Bangla (BdSL) & 4,000 & Open & 27.3 & --- \\
\midrule
\textbf{Ours} & \textbf{Bangla (BdSL)} & \textbf{4,831} & \textbf{Open} & \textbf{51.7} & \textbf{61.4} \\
\bottomrule
\end{tabular}%
}
\end{table*}

\subsection{Discussion}
\label{sec:discussion}
The results show that monolingual Bangla pre-training and data augmentation can improve bidirectional gloss-sentence translation. However, IsharaBART does not outperform all baselines on every metric. On G2S, BanglaT5 achieves higher BLEU-2, BLEU-3, and BLEU-4 scores. After augmentation, IsharaBART reaches a similar BLEU-4 score (51.74 versus 52.24). However, IsharaBART uses 56M parameters, while BanglaT5 uses 247M. On S2G, IsharaBART achieves the highest BLEU-4 score among the evaluated models in both data settings. With augmented data, it reaches 61.40 BLEU-4. The score for mBART-50 is 60.82, despite its 611M parameters. These results show that IsharaBART offers a good balance between model size and translation performance.

The pre-training ablation provides further insight. Pre-training gives the largest relative gains when the model has only 1,000 annotated pairs. These gains remain after data augmentation. Pre-training also reduces the number of epochs needed to reach the best validation BLEU-4 score. These results suggest that monolingual Bangla pre-training helps translation when labeled data are limited.

Data augmentation also improves performance, but the gains differ by translation direction. For IsharaBART, BLEU-4 increases by 8.44 points on G2S and 22.01 points on S2G. One possible reason is that S2G requires the model to generate gloss sequences from natural Bangla sentences. The model must learn specific output patterns for this task. The annotated dataset may not provide enough examples of these patterns. Augmentation may help the model learn them. However, this explanation remains a hypothesis. The experiments do not identify the exact linguistic features behind the larger S2G gain.

This study extends previous BdSL research by evaluating both G2S and S2G translation in the same experimental framework. It also examines monolingual pre-training and targeted data augmentation. The results show that the two directions can produce different outcomes. After augmentation, IsharaBART comes close to BanglaT5 on G2S. It achieves the highest S2G BLEU-4 score among the evaluated models. These differences show why studies should evaluate both directions. Results from one direction may not reflect performance on the other. Model selection should also consider the target task and available computing resources.

Human evaluation provides further insight into translation quality. The mean adequacy scores are 88.8 for G2S and 87.8 for S2G. However, only 66\% of G2S outputs and 56\% of S2G outputs meet both adequacy and fluency criteria. This suggests that the model often preserves the intended meaning. However, producing fluent and consistent translations remains difficult. Inter-rater agreement was modest to moderate. The evaluation also included only 50 outputs per direction. These results show promise, but they do not establish reliable translation in every case.

The LoRA experiments show another trade-off. LoRA reduces the number of trainable parameters by 99.3\%. However, BLEU-4 decreases by 28.0--52.2\% compared with full fine-tuning. The decline is greater for S2G than for G2S under the tested configuration. This suggests that LoRA does not fully match full fine-tuning in these experiments.

\section{Conclusion and Future Work}
\label{sec:conclusion}
In this study, we presented IsharaBART, a bidirectional translation model for BdSL. It translates between gloss sequences and Bangla sentences in both directions. Our results show that monolingual pre-training and data augmentation improve translation quality. The model also reduces training time by 62.5\% to 90.4\%. It maintains good structural, semantic, and lexical consistency. We compared IsharaBART with four baseline models under the same conditions. The results show that our approach achieves comparable or better translation quality. Its small size and faster training may support accessibility tools for education, healthcare, and daily communication.

However, this study has some limitations. The annotated and augmented dataset is small compared to high-resource benchmarks. This may limit the model's ability to handle rare sentence structures and glosses. The model also depends on accurate gloss sequences from a vision-based sign recognition system. We did not study this component. Recognition errors may occur due to poor lighting or differences between signers. Our monolingual approach may also miss benefits from multilingual training.

IsharaBART has a compact architecture and needs fewer training epochs. These features may help on devices with limited resources. However, we did not measure inference time, memory use, or energy consumption. We also did not test the model on mobile or edge devices. Its efficiency in real-world settings remains unclear.

The model translates gloss sequences and Bangla text. It does not process sign-language videos directly or generate visual signs. A complete translation system would need a sign recognition model and a method to generate sign-language output. IsharaBART could serve as the text translation component of such a system.

The approach may also work for other low-resource sign languages, but we have not tested this. Future work should expand the dataset and include more signers and regional gloss variations. Larger and more diverse groups of human evaluators should assess the translations. Human evaluations of the baseline models would also show whether the automatic scores reflect real improvements. Tests of inference speed, memory use, and energy consumption are also needed. Integrating video-based sign recognition would help assess the model in real-world communication systems.

Future work should expand the dataset through annotations from the deaf community. We should also explore end-to-end video-to-text models that do not depend on gloss inputs. Hybrid multilingual and monolingual pre-training may help the model handle linguistic variation. Another direction is to test span-corruption and hybrid denoising objectives. These methods may help close the remaining BLEU-3 and BLEU-4 gaps in G2S translation, especially compared with BanglaT5. The goal is to improve translation quality without increasing model size or training time. Finally, real-world deployment studies could help measure the model's practical value and guide further improvements for other low-resource sign languages.

\section*{Disclosure statement}
The author(s) declare(s) no potential conflict of interest regarding the publication of this paper.

\section*{Acknowledgment}
This work was supported by the ICT fellowship (FY 2024–2025) under grant code 56.00.0000.052.33.0003.24–115 provided by the Information and Communication Technology Division, Ministry of Posts, Telecommunications and Information Technology, Government of the People’s Republic of Bangladesh.

 \bibliographystyle{elsarticle-num} 
\bibliography{bibliography}

\appendix
\section{Training Configuration and Hyperparameters}
\label{ap:training_config}
We provide complete training configurations for IsharaBART pre-training and fine-tuning experiments to ensure reproducibility. All models were trained using the Hugging Face Transformers library with PyTorch backend.

\subsection{IsharaBART Pre-training}

IsharaBART was pre-trained on a monolingual Bangla corpus using a denoising autoencoder objective with custom corruption strategies. Table~\ref{tab:pretrain_config} summarizes the key hyperparameters. We use polynomial learning rate decay as recommended in the original BART paper, with a relatively high initial learning rate ($5 \times 10^{-4}$) to accommodate the smaller model size and focused pre-training corpus. Training was conducted in bfloat16 precision for numerical stability and computational efficiency. The effective batch size of 256 (64 per device $\times$ 4 gradient accumulation steps) balances training stability with memory constraints of encoder--decoder architectures.

\subsection{Fine-tuning Configuration}

All models (IsharaBART and baselines) were fine-tuned on the gloss translation tasks using identical or carefully matched configurations to ensure fair comparison. Table~\ref{tab:finetune_config} presents the hyperparameters used for each model and Tale \ref{tab:lora-config} present the LoRA configuration for parameter-efficient fine-tuning of IsharaBART. Configuration differences across models reflect preliminary validation experiments and model-specific characteristics: mT5-Small benefits from a higher learning rate ($1 \times 10^{-3}$) due to its architecture and pre-training objective, while BanglaGPT requires a lower learning rate ($1 \times 10^{-5}$) and larger gradient accumulation to stabilize training in the decoder-only setting.

For all encoder--decoder models (IsharaBART, mBART-50, mT5-Small, BanglaT5), we use beam search with 4 beams during generation for evaluation. Early stopping with patience of 3--5 epochs prevents overfitting on the low-resource fine-tuning datasets. Model selection is based on validation BLEU-4 for encoder--decoder models and validation loss for BanglaGPT, reflecting the different training objectives and generation strategies.

\begin{table}[t]
\centering
\caption{IsharaBART pre-training configuration. All training used AdamW optimizer with $\beta_1=0.9$, $\beta_2=0.999$, $\epsilon=10^{-8}$.}
\label{tab:pretrain_config}
\begin{tabular}{l|l}
\toprule
\textbf{Hyperparameter} & \textbf{Value} \\
\midrule
Training Steps & 30,000 \\
Batch Size (per device) & 64 \\
Gradient Accumulation & 4 steps \\
Effective Batch Size & 256 \\
\midrule
Learning Rate & $5 \times 10^{-4}$ \\
LR Schedule & Polynomial decay \\
Warmup Steps & 3,000 \\
Weight Decay & 0.1 \\
Max Gradient Norm & 1.0 \\
\midrule
Precision & bfloat16 \\
Gradient Checkpointing & False \\
Random Seed & 42 \\
\midrule
Evaluation Strategy & Every 1,000 steps \\
Save Strategy & Every 5,000 steps \\
Model Selection & Best validation loss \\
\bottomrule
\end{tabular}
\end{table}

\begin{table}[t]
\centering
\caption{Fine-tuning configuration for all models. ES = Early Stopping patience (epochs). All models use AdamW optimizer except where noted.}
\label{tab:finetune_config}
\resizebox{\textwidth}{!}{%
\begin{tabular}{l|ccccc}
\toprule
\textbf{Hyperparameter} & \textbf{IsharaBART} & \textbf{mBART-50} & \textbf{mT5-Small} & \textbf{BanglaT5} & \textbf{BanglaGPT} \\
\midrule
Epochs & 20 & 20 & 20 & 20 & 20 \\
Batch Size (per device) & 16 & 16 & 16 & 16 & 16 \\
Gradient Accumulation & 2 & 2 & 2 & 2 & 4 \\
Effective Batch Size & 32 & 32 & 32 & 32 & 64 \\
\midrule
Learning Rate & $1 \times 10^{-4}$ & $1 \times 10^{-4}$ & $1 \times 10^{-3}$ & $1 \times 10^{-4}$ & $1 \times 10^{-5}$ \\
LR Schedule & Linear decay & Linear decay & Linear decay & Linear decay & Linear decay \\
Weight Decay & 0.01 & 0.01 & 0.01 & 0.01 & 0.01 \\
Max Gradient Norm & 1.0 & 1.0 & 1.0 & 1.0 & 1.0 \\
\midrule
Precision & fp32 & fp32 & fp32 & fp32 & fp32 \\
Beam Size & 4 & 4 & 4 & 4 & -- \\
Generation Max Length & 128 & 128 & 128 & 128 & 128 \\
\midrule
ES Patience & 5 & 5 & 5 & 5 & 5 \\
Model Selection Metric & BLEU-4 & BLEU-4 & BLEU-4 & BLEU-4 & Validation loss \\
\bottomrule
\end{tabular}%
}
\end{table}

\begin{table}[t]
\centering
\small
\begin{tabular}{l l}
\toprule
Hyperparameter & Value \\
\midrule
Task type & Seq2Seq language modeling \\
LoRA rank ($r$) & 8 \\
LoRA scaling factor ($\alpha$) & 16 \\
LoRA dropout & 0.1 \\
Target modules & q\_proj, k\_proj, v\_proj, out\_proj \\
\bottomrule
\end{tabular}
\caption{LoRA configuration used for parameter-efficient fine-tuning.}
\label{tab:lora-config}
\end{table}

All experiments were conducted using Hugging Face Transformers (v4.57) and PyTorch 2.9 on NVIDIA A100 GPU for pre-training and Tesla T4 GPU for fine-tuning. For pre-training, we implemented a custom data collator that applies BART-style corruption including text infilling, sentence permutation, and token masking, with special handling for Bangla punctuation tokens (danda and question marks). Fine-tuning used the standard \texttt{DataCollatorForSeq2Seq} for dynamic padding.

\end{document}